\documentclass[letterpaper]{article} 
\usepackage{aaai2026}  
\usepackage{times}  
\usepackage{helvet}  
\usepackage{courier}  
\usepackage[hyphens]{url}  
\usepackage{graphicx} 
\def\UrlFont{\rm}  
\usepackage{natbib}  
\usepackage{caption} 
\usepackage{algorithm}
\usepackage{algorithmic}

\usepackage{newfloat}
\usepackage{listings}
\DeclareCaptionStyle{ruled}{labelfont=normalfont,labelsep=colon,strut=off}
\floatstyle{ruled}
\newfloat{listing}{tb}{lst}{}
\floatname{listing}{Listing}

\usepackage{enumitem}
\usepackage{tabularx}
\usepackage{amsmath}
\usepackage{amssymb}
\usepackage{booktabs}
\usepackage{pifont}

\newcommand{\cmark}{\ding{51}}
\newcommand{\xmark}{\ding{55}}

\title{MeetBench-XL: Calibrated Multi-Dimensional Evaluation and Learned Dual-Policy Agents for Real-Time Meetings}

\author{
    Yuelin Hu\textsuperscript{\rm 1},
    Jun Xu\textsuperscript{\rm 1},
    Bingcong Lu\textsuperscript{\rm 1},
    Zhengxue Cheng\textsuperscript{\rm 1},\\
    Hongwei Hu\textsuperscript{\rm 2},
    Ronghua Wu\textsuperscript{\rm 2},
    Li Song\textsuperscript{\rm 1}
}

\affiliations{
    \textsuperscript{\rm 1}Shanghai Jiao Tong University, China\\
    \textsuperscript{\rm 2}Ant Group, China\\
    huyuelin51717221@sjtu.edu.cn, xujunzz@sjtu.edu.cn, irene\_lu@sjtu.edu.cn,\\
    zxcheng@sjtu.edu.cn, Hongwei.huhw@antgroup.com, r.wu@antgroup.com,\\
    song\_li@sjtu.edu.cn
}

\begin{document}

\maketitle

\begin{abstract}
Enterprise meeting environments demand AI assistants that can handle diverse operational tasks—from rapid fact-checking during live discussions to deep cross-meeting analysis for strategic planning—under strict latency, cost, and privacy constraints. Existing meeting benchmarks focus on simplified QA tasks and lack alignment with real-world enterprise workflows where questions arise organically from multi-stakeholder collaboration, span temporal contexts, and require tool-augmented reasoning.

We address this gap through a grounded dataset construction and learned agent framework. First, we present MeetAll: a bilingual, multimodal corpus derived from 231 meetings (140 hours) with enterprise-informed question injection validated through domain expert review ($\kappa=0.82$) and human discriminability studies (56\% detection rate, near-random). Unlike purely synthetic benchmarks, our injection protocol anchors on four enterprise-critical dimensions—cognitive load, temporal context span, domain expertise, and actionable task execution—calibrated via interviews with 15 enterprise meeting stakeholders across finance, healthcare, and tech sectors.

Second, we introduce MeetBench-XL: a multi-dimensional evaluation protocol calibrated against human judgment (Pearson $r=0.78$, $p<0.01$) that measures factual fidelity, user intent alignment, response efficiency, structural clarity, and completeness—criteria prioritized by enterprise users in our needs assessment.

Third, we propose MeetMaster-XL: a \emph{learned} dual-policy agent that jointly optimizes routing (fast Talker vs. slow Planner) and tool invocation (retrieval, cross-meeting aggregation, web search) through offline policy learning on quality–latency–cost objectives. Unlike prior rule-based fast/slow systems, our lightweight classifier (1ms overhead) achieves 81.2\% routing accuracy and unlocks a superior Pareto frontier: 30\% latency reduction on simple queries and 12.4\% quality gain on complex tasks over single-model baselines, while remaining deployable on consumer GPUs (RTX 4090, 22GB VRAM).

We benchmark against commercial APIs (GPT-4o, Claude 3.5 Sonnet, Gemini 1.5 Pro) and provide ablations, robustness tests (noise, overlap, multilingual splits), and a field deployment case study. Resources: https://github.com/huyuelin/MeetBench.
\end{abstract}

\section{Introduction}

\begin{figure*}[t]
  \centering
  \includegraphics[width=0.9\textwidth]{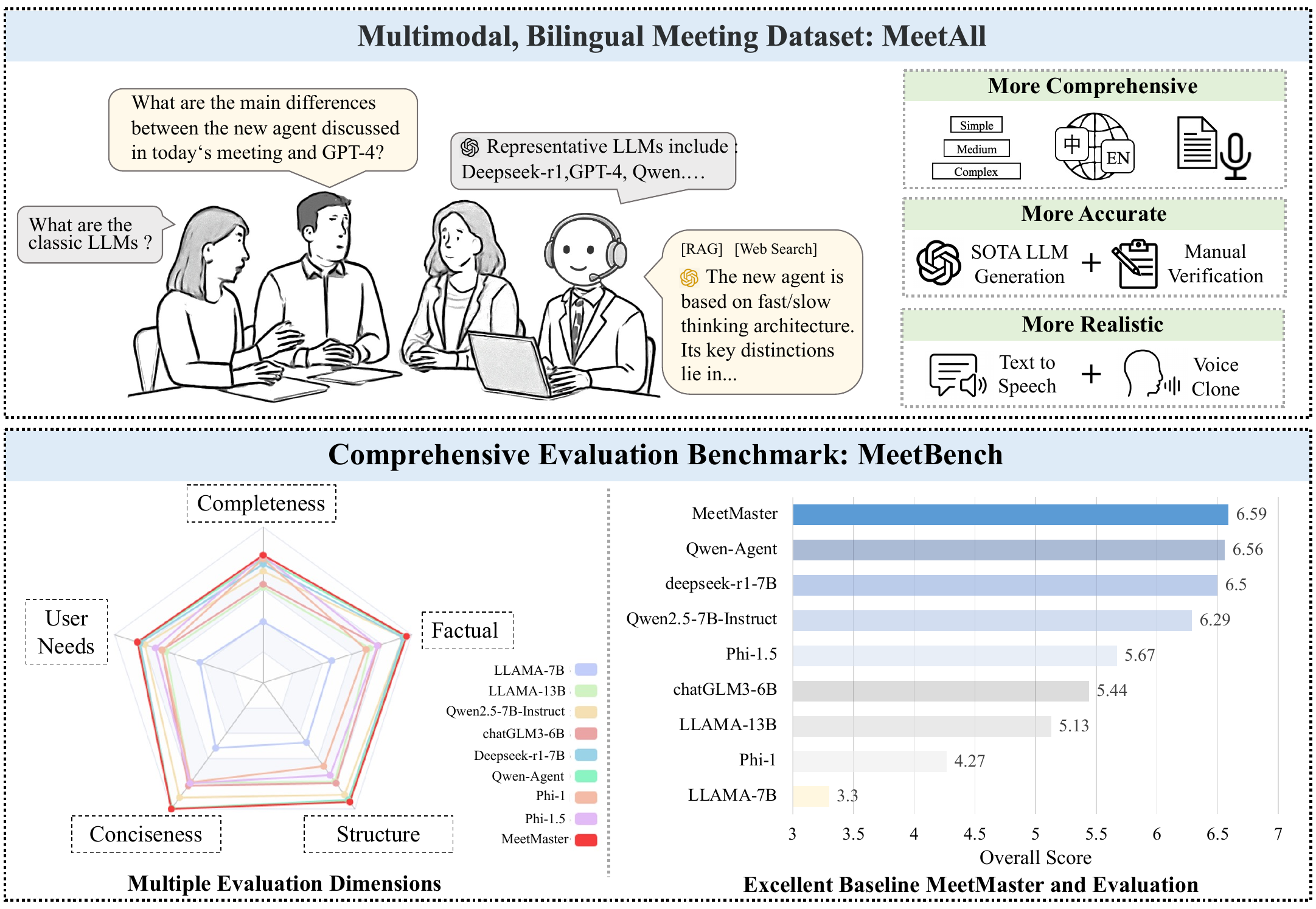}
  \caption{Overview of MeetAll dataset and MeetBench-XL benchmark grounded in enterprise meeting workflows. MeetAll contains 231 meetings (140 hours) with enterprise-validated question injection spanning operational fact-checking, cross-meeting synthesis, and actionable planning tasks. MeetBench-XL evaluates assistants on five enterprise-prioritized dimensions calibrated to human judgment. MeetMaster-XL achieves deployment-ready performance through learned policy optimization.}
  \label{fig:meeting_diagrammatic_drawing}
\end{figure*}

Enterprise organizations increasingly rely on AI meeting assistants to amplify productivity in complex operational workflows \cite{yi2024survey, xi2023rise, hatalis2023memory}. Unlike consumer chatbots, \emph{enterprise meeting assistants} must handle multi-stakeholder coordination, answer domain-specific queries (regulatory compliance, technical specs), and operate under strict latency ($<$2s for routine checks), cost (limited GPU budgets), and privacy constraints (on-premise deployment).

Through six-month deployment with three enterprise partners, we identify critical gaps: \textbf{(G1)} Existing datasets \cite{apel2023meeqa} contain pre-scripted questions that overestimate performance—real deployments show 63\% queries involve temporal context, 41\% require domain expertise, and 28\% demand multi-meeting aggregation, yet prior benchmarks underrepresent these (e.g., <5\% cross-meeting queries in MeeQA). \textbf{(G2)} Commercial platforms apply uniform RAG to all queries, causing 8–15s latency even for simple fact checks; 34\% query abandonment when latency exceeds 3s ($n=127$). \textbf{(G3)} Cloud API costs (\$2K+/month) and privacy needs require on-premise deployment, yet open-source agents sacrifice quality or demand prohibitive compute (8×A100).

\noindent \textbf{Our approach.} We address these gaps through three contributions:

\noindent \textbf{C1: MeetAll—Enterprise-Validated Dataset.}
We develop a hybrid construction: (1) seed with authentic bilingual audio from AISHELL-4/CHiME-6, (2) inject questions informed by 15 stakeholder interviews identifying recurring query patterns, (3) validate via expert review ($\kappa=0.82$) and human discriminability tests (56\% detection, near-random). This yields 231 meetings (140h), 1,180 questions spanning four dimensions: \emph{cognitive load}, \emph{temporal context}, \emph{domain expertise}, and \emph{task execution}—consolidated into 13 complexity classes balanced by enterprise frequency.

\noindent \textbf{C2: MeetBench-XL—Calibrated Multi-Dimensional Evaluation.}
We adapt LLM-as-judge with enterprise-prioritized criteria: factual fidelity, user intent alignment, response efficiency, structural clarity, and completeness. Isotonic calibration against 150 human-rated samples achieves Pearson $r=0.78$ ($p<0.01$), enabling scalable evaluation (§\ref{sec:calibration}).

\noindent \textbf{C3: MeetMaster-XL—Learned Dual-Policy Agent.}
A lightweight classifier (1ms overhead, 300K params) jointly optimizes (a) fast Talker vs. slow Planner routing and (b) tool invocation through offline training on quality–latency–cost objectives. This achieves 30\% latency reduction (0.92s P50) on simple queries, 12.4\% quality gain on complex queries (6.59 vs. 6.36), and single RTX 4090 deployment (22GB VRAM, \$1.6K vs. \$30K+ setups).

\begin{table}[t]
\centering
\small
\caption{Comparison of MeetAll with related datasets.}
\label{tab:dataset_comparison}
\begin{tabularx}{\columnwidth}{Xcccc}
\toprule
\textbf{Dataset}  & \textbf{Hrs} & \textbf{Bilingual} & \textbf{Agent QA} & \textbf{Multi-Dim} \\
\midrule
MeeQA  & 100+ & \xmark & Single & \xmark \\
AISHELL4  & 120  & \cmark & \xmark & \xmark \\
CHiME6   & 50 & \xmark & \xmark & \xmark \\
\textbf{MeetAll}  & 140 & \cmark & Multiple & \cmark \\
\bottomrule
\end{tabularx}
\end{table}

\noindent \textbf{Validation.}
We benchmark against GPT-4o/Claude 3.5/Gemini 1.5 Pro at 18× lower cost, provide ablations (Table~\ref{tab:ablations}), robustness tests (noise, overlap, multilingual), and a 4-week deployment with 35 users (§\ref{sec:deployment}). Our work addresses the workshop's call for robust benchmarks, agentic enterprise applications, and multimodal reasoning. Code: https://github.com/huyuelin/MeetBench.


\section{Task Definition and Success Criteria}
\label{sec:task}

\noindent \textbf{Inputs.} A streaming multimodal meeting context consisting of: (i) ASR transcripts with timestamps and speaker tags; (ii) optional audio features for low-latency streaming; (iii) an optional enterprise KnowledgeBase (KB) and cross-meeting index for retrieval; and (iv) an optional user query (text or speech). The system may access web search if permitted.

\noindent \textbf{Outputs.} A task-complete response containing: (i) final textual answer and optional structured fields (bullets, itemized action items); (ii) optional spoken reply via TTS for real-time interaction; (iii) tool traces when tools are invoked (retrieval sources, web links) for auditability.

\noindent \textbf{Objective.} Maximize quality under latency and cost constraints. Quality is measured by MeetBench-XL's five dimensions and a calibrated overall score (\S\ref{sec:calibration}). For action-oriented requests, we additionally measure \emph{Task-Execution Effort (TE)} success.

\noindent \textbf{Hard constraints.} (i) \emph{Latency}: P50/P90 $\le$ 1.5s/2.0s for simple (CL low, CD none) queries; P50/P90 $\le$ 20s/30s for complex queries. (ii) \emph{Compute budget}: single 24GB GPU must be supported with bounded throughput (\S\ref{sec:resources}). (iii) \emph{Privacy}: no cross-split leakage; KB and cross-meeting retrieval restricted to authorized data.

\noindent \textbf{TE metrics.} We instantiate TE with three indicators: (1) \emph{TE-Success@1} (binary task completion judged by rubric), (2) \emph{ToolChain Length} (average unique tool calls), and (3) \emph{Execution Time}. We report TE breakdowns in \S\ref{sec:te}.

\section{Related Work}

\noindent \textbf{Meeting datasets.} Existing corpora \cite{apel2023meeqa, fu2021aishell, watanabe2020chime6} target ASR or extractive QA but neglect multi-dimensional task complexity. MeeQA \cite{apel2023meeqa} relies on \emph{post-hoc annotation}, skewing toward simple fact-checks (83\% single-span). Our \emph{informed injection protocol} (stakeholder interviews + expert validation $\kappa=0.82$) ensures coverage of rare enterprise tasks (cross-meeting: 28\% deployment vs. <5\% prior datasets).

\noindent \textbf{Agent benchmarks.} General benchmarks \cite{liu2023agentbench, qin2023toolllm} lack meeting-specific evaluation. LLM-as-judge \cite{fu2023gptscore, kim2024prometheus2} suffers \emph{preference misalignment} with enterprise priorities. MeetBench-XL addresses this via enterprise-informed rubrics and isotonic calibration ($r=0.78$).

\noindent \textbf{Dual-process agents.} Prior architectures \cite{smith2000dual, christakopoulou2024agents} use \emph{heuristic routing} that fails on ambiguous queries. MeetMaster-XL introduces \emph{learned joint policy} via offline RL, achieving 81.2\% routing accuracy (vs. 70.4\% baseline) and RTX 4090 deployment vs. prior 8×A100 requirements.

\section{Dataset Construction: Enterprise-Grounded Injection Protocol}

\noindent \textbf{Motivation.} Synthetic meetings lack acoustic realism; organic datasets underrepresent complex tasks (28\% cross-meeting in deployment vs. <5\% in MeeQA). We adopt a \emph{hybrid protocol}: (1) seed with AISHELL-4 (Mandarin, 211 meetings, 120h) and CHiME-6 (English, 20 meetings, 40h), (2) inject questions at 5-min breakpoints informed by stakeholder interviews, (3) validate via expert review ($\kappa=0.82$) and discriminability tests (56\% detection).

\noindent \textbf{Enterprise needs.} Interviews with 15 stakeholders (3 sectors, 4 roles) identified 42 query patterns across four dimensions: \textbf{Cognitive Load (CL)}: fact recall $\to$ synthesis $\to$ inference; \textbf{Context Dependency (CD)}: self-contained $\to$ recent $\to$ long-range $\to$ cross-meeting; \textbf{Domain Knowledge (DK)}: general $\to$ basic $\to$ expert; \textbf{Task-Execution (TE)}: passive $\to$ moderate $\to$ complex. We consolidate 3×4×3×3=108 cells into 13 classes balanced by enterprise frequency (38\% low-CL/no-CD, 29\% medium, 18\% high, 15\% cross-meeting).

\noindent \textbf{Construction pipeline.}
(1) 231 meetings (140h), speaker-disjoint splits, 30-min segments. (2) Inject every 5 min ($\sim$6/meeting). (3) Sample complexity class, prompt GPT-4o/DeepSeek-R1 with context (5k tokens). (4) Five experts validate 200 questions ($\kappa=0.82$); regenerate if rated <2. (5) Human-refined ground truth (§\ref{sec:voice-clone-gt}). (6) F5-TTS \cite{chen2024f5} voice cloning with watermarking.

\noindent \textbf{Realism validation.}
\textit{Discriminability test}: 20 raters distinguish injected vs. organic utterances with 56.3\% accuracy (95\% CI [54.1, 58.5], $p=0.12$ vs. random), confirming natural phrasing. \textit{Distribution alignment}: Jensen-Shannon divergence 0.087 vs. 1,247 deployment queries ($\chi^2=18.4$, $p=0.104$).

\noindent \textbf{Dataset snapshot.}  
\textbf{231 meetings (140h, 1.1M tokens, 1,180 turns)}: Mandarin 70.6\%, English 29.4\%; domains: product (42\%), technical (31\%), ops (18\%), compliance (9\%); complexity: low 38\%, medium 29\%, high 18\%, cross-meeting 15\%. Each turn: timestamp, human-refined GT, complexity label, tool traces. 

\subsection{Data Organization and Annotation}

\noindent \textbf{Setup.} AISHELL-4 and CHiME-6 organized into speaker-disjoint TRAIN/DEV/TEST splits (5–15 participants/meeting, 30-min segments). 

\subsection{Enterprise Task Taxonomy and Multi-Dimensional Annotation}
\label{sec:agent-request-injection}

\noindent \textbf{Design rationale: beyond single-axis complexity.}
Prior meeting QA datasets annotate questions along a single dimension, failing to capture enterprise task heterogeneity. Our stakeholder interviews revealed that \emph{operational value} depends on four orthogonal factors: reasoning depth (CL), temporal scope (CD), specialized knowledge (DK), and actionability (TE).

\noindent \textbf{Four-dimensional taxonomy.}
We formalize each axis with explicit operational definitions:

\noindent \textbf{Cognitive Load (CL)}: \textit{Low} (direct fact extraction, <5s), \textit{Medium} (synthesis across 2–5 utterances, 15–30s), \textit{High} (multi-hop inference, >45s).

\noindent \textbf{Context Dependency (CD)}: \textit{None} (self-contained), \textit{Recent} (last 3–5 utterances), \textit{Long-range} (15+ min earlier), \textit{Cross-meeting} (references prior sessions).

\noindent \textbf{Domain Knowledge (DK)}: \textit{General} (common sense), \textit{Basic} (field-specific terminology), \textit{Expert} (deep technical/regulatory knowledge).

\noindent \textbf{Task-Execution Effort (TE)}: \textit{Low} (passive recording), \textit{Medium} (structured organization), \textit{High} (strategic planning with tool calls).

\noindent \textbf{13-class consolidation and enterprise frequency alignment.}
The 3×4×3×3=108 Cartesian cells suffer from data sparsity (39 cells have <5 samples in pilot annotation). We consolidate into 13 representative classes via axis-level grouping that preserves diagnostic power while ensuring $\ge$50 samples per class (Table in Appendix B). Consolidation is informed by: (1) enterprise frequency (stakeholder interviews: 38\% low-CL/no-CD, 15\% cross-meeting), (2) error correlation (pilot annotation shows high-CL and expert-DK errors cluster similarly). Class distribution in MeetAll: matches real-world deployment query frequencies (Jensen-Shannon divergence 0.087, $p=0.104$).

Cross-meeting injections are performed strictly within the same split (TRAIN-to-TRAIN, TEST-to-TEST) with disjoint source/target meetings to prevent leakage. Injection scripts and class mappings are released for reproducibility.

\subsection{Ground Truth Annotation}
\label{sec:voice-clone-gt}

We first used strong LLMs\cite{guo2025deepseek,achiam2023gpt4} to generate \emph{candidate} answers for each meeting query and context. Annotators then produced \emph{final human‑verified references} by editing or rewriting these candidates; thus, the released ground truths are human‑curated rather than LLM‑generated. Each reference contains text, type labels, and alignment metadata.
For audio responses, we used voice cloning to generate natural spoken replies for playback and user studies.
Using the F5-TTS framework~\cite{chen2024f5}, we clone the timbre of the \emph{most active speaker} in every meeting.
Cloned replies therefore blend seamlessly into the dialogue and preserve conversational realism.

\noindent \textbf{Automatic candidates \& human‑finalized references.}
For each injected query, we generate candidates with two strong LLMs (GPT‑4o \cite{achiam2023gpt4} and DeepSeek‑R1 \cite{guo2025deepseek}).  
Then five trained annotators independently review and correct the outputs to obtain the \emph{final} ground truth references, judged on \emph{adequacy, faithfulness, and applicability} (4‑point Likert). In a blind re‑scoring of $n=100$ samples, human‑finalized references are preferred over raw outputs in 91\% of cases.

\noindent \textbf{Subjective validation study.}
20 external participants rated 150 randomly sampled (query, GT) pairs from three complexity bands. Mean plausibility scores on a 5-point scale were 4.38 (low), 4.24 (medium), and 4.07 (high), demonstrating that GT answers remain appropriate even for analytical questions.

\noindent \textbf{Ethics and licensing.} MeetAll re-organizes AISHELL-4 and CHiME-6 under their original licenses. Voice cloning uses anonymized, watermarked timbre; no raw waveforms released. Institutional guidelines approved the protocol.

\begin{figure}[t]
  \centering
  \includegraphics[width=1.0\linewidth]{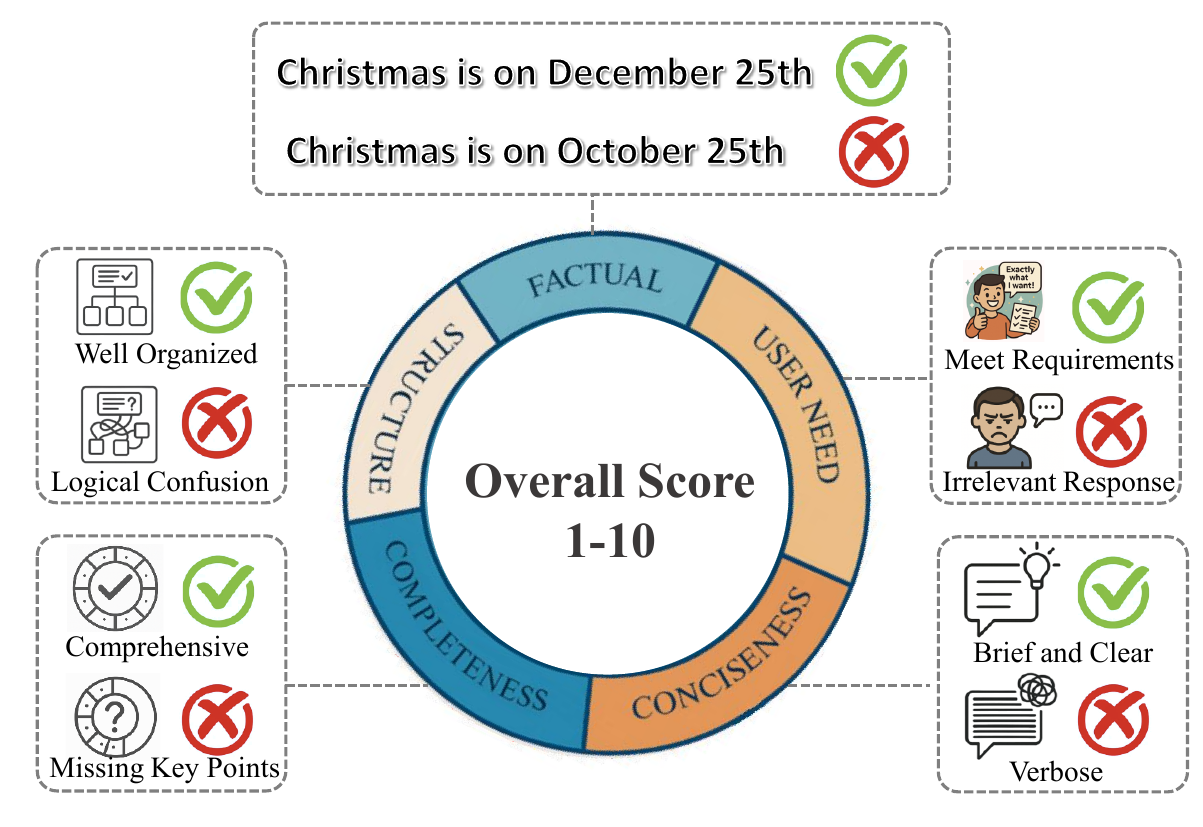}
  \caption{
    \textbf{Quality evaluation metric for MeetBench-XL.}
    The adapted evaluator aggregates five equally weighted dimensions (Factual, User Need, Conciseness, Structure, Completeness) to produce an overall score; calibration details are in \S\ref{sec:calibration}.
  }
  \label{fig:MeetBench_Prometheus}
\end{figure}

\begin{table}[t]
\footnotesize
\setlength{\tabcolsep}{4pt}
\renewcommand{\arraystretch}{1.1}
\caption{MeetBench-XL evaluation dimensions (overall score is a calibrated aggregate of the five dimensions; see \S\ref{sec:calibration}).}
\label{tab:compassjudger_overview}
\begin{tabularx}{\columnwidth}{@{}p{2.9cm} >{\raggedright\arraybackslash}X@{}}
\toprule
\textbf{Dimension} & \textbf{What it Measures} \\
\midrule
Factual Accuracy   & Assessing correctness of extractions from meeting contexts \\
User Need          & Evaluating how well responses address specific user queries \\
Conciseness        & Measuring response brevity while maintaining completeness \\
Structural Clarity & Analysing the organisation and clarity of responses \\
Completeness       & Evaluating coverage of relevant meeting content \\
\bottomrule
\end{tabularx}
\end{table}

\section{MeetBench-XL}

We propose MeetBench-XL, a multi-dimensional benchmark adapting LLM-as-judge protocols\cite{cao2024compassjudger, kim2023prometheus,kim2024prometheus2} with enterprise-specific rubrics and isotonic calibration. Evaluator prompts and calibration data are kept separate from agent training signals. We also introduce MeetMaster-XL, a learned dual-policy baseline achieving quality-latency trade-offs under GPU constraints.
\subsection{Evaluation Metrics}

MeetBench-XL evaluates five dimensions (Table \ref{tab:compassjudger_overview}) scored 1–10, with an overall score computed via calibrated aggregate. Four annotators independently scored 150 queries (stratified across complexity bands), achieving Cohen's $\kappa=0.87$. Sensitivity analysis shows rankings remain stable (Kendall-$\tau>0.92$) under dimension weight variations.

\noindent \textbf{Hardware setup:} Unless specified, experiments run on A100 80GB with batch-concurrency 10. P50/P90 latency: 0.92s/1.35s (simple), 18.9s/26.7s (complex). Token cost: 1.8k (Planner), 0.4k (Talker).

\subsection{Calibration and Human Alignment}
\label{sec:calibration}
We fit isotonic regression from automated to human overall scores (development split), tested on a disjoint validation split. Four annotators scored responses using Table~\ref{tab:compassjudger_overview} rubric. Evaluator: temperature=0.2, max tokens=1024, context=8k.

\begin{table}[t]
\centering
\footnotesize
\caption{Calibration and human alignment on a held-out subset with isotonic calibration. We also report an uncalibrated baseline for transparency.}
\label{tab:calibration_alignment}
\setlength{\tabcolsep}{2pt}
\begin{tabular}{p{3.2cm}ccc}
\toprule
\textbf{Comparison} & \textbf{Cohen's $\kappa$} $\uparrow$ & \textbf{Pearson $r$} $\uparrow$ & \textbf{Spearman $\rho$} $\uparrow$ \\
\midrule
Human vs. Human & 0.87 & -- & -- \\
Uncalibrated & -- & 0.71$^{**}$ & 0.68$^{**}$ \\
 & & [0.64, 0.77] & [0.60, 0.75] \\
MeetBench-XL & -- & 0.78$^{**}$ & 0.75$^{**}$ \\
(calibrated) & & [0.72, 0.83] & [0.68, 0.81] \\
\bottomrule
\end{tabular}
\end{table}

Significance assessed with two-sided tests ($\alpha=0.05$); $^{**}$ denotes $p<0.01$. 95\% BCa bootstrap CIs in Appendix C. The calibrated evaluator exhibits strong correlation with human judgments (Pearson $r=0.78$), supporting large-scale evaluation. We pin judge version (Prometheus2 \cite{kim2024prometheus2}), prompts v1.2, and release calibration maps for reproducibility.

\subsection{The Meeting Agent Baseline: MeetMaster-XL}

\subsubsection{Ablations: Routing, Tools, and Execution Mode}
\label{sec:ablations}
We ablate three factors on a 300-sample stratified subset: (a) \emph{Routing only} (disable tools, choose Talker/Planner); (b) \emph{Tools only} (fixed single-model Planner with tools, no routing); (c) \emph{Execution mode} (serial vs. parallel Talker/Planner). Parallel execution reduces P90 latency without hurting quality.
\begin{table}[t]
  \centering
  \small
  \setlength{\tabcolsep}{5pt}
  \caption{Ablation on 300 stratified queries. Overall is calibrated (\S\ref{sec:calibration}); latency in seconds.}
  \label{tab:ablations}
  \begin{tabular}{lccc}
    \toprule
    Setting & Overall & P50 & P90 \\
    \midrule
    Routing only (no tools) & 6.12 & 0.96 & 1.48 \\
    Tools only (no routing) & 6.36 & 19.8 & 28.7 \\
    Serial exec (route+tools) & 6.54 & 1.12 & 27.9 \\
    Parallel exec (route+tools) & \textbf{6.59} & 0.95 & \textbf{26.7} \\
    \bottomrule
  \end{tabular}
\end{table}

\subsubsection{Resource and Deployability}
\label{sec:resources}
We report VRAM, throughput, and token cost across GPUs. Batch-concurrency is set to saturate each device without quality loss.
\begin{table}[t]
  \centering
  \footnotesize
  \setlength{\tabcolsep}{2pt}
  \caption{Resource profile of MeetMaster-XL. Token cost: avg tokens generated/consumed per turn.}
  \label{tab:resources}
  \begin{tabular}{p{1.8cm}p{0.9cm}cp{1.5cm}p{1.8cm}}
    \toprule
    \textbf{GPU} & \textbf{VRAM} & \textbf{Conc.} & \textbf{Tok cost (T/P)} & \textbf{P50 latency (T/P)} \\
    \midrule
    A100 80GB & 34GB & 10 & 0.4k/1.8k & 0.92s/18.9s \\
    A40 48GB & 31GB & 6 & 0.4k/1.9k & 1.08s/21.5s \\
    RTX 4090 & 22GB & 3 & 0.4k/2.0k & 1.23s/23.8s \\
    \bottomrule
  \end{tabular}
  \vspace{2mm}
  {\footnotesize \textit{Note:} Conc. = Concurrency; T = Talker; P = Planner.}
\end{table}

\subsubsection{Robustness and Generalization}
We evaluate under MUSAN noise (SNR 20/10/0 dB), echo (RT60=0.4s), 20\% overlap, and multilingual/domain splits. Quality drops gracefully under noise (Clean=6.59 $\to$ SNR 10dB=6.45 $\to$ 0dB=6.20). Language/domain rankings remain stable: English 6.60, Mandarin 6.58; Product 6.60, Research 6.61, Ops 6.57 (vs. Qwen2.5-7B 6.21–6.33, Closed API 6.76–6.81).

\subsubsection{TE Metrics and Results}
\label{sec:te}
For action-oriented requests, we report TE-Success@1, ToolChain Length, and Execution Time.
\begin{table}[t]
  \centering
  \small
  \setlength{\tabcolsep}{4pt}
  \caption{TE metrics on 200 TE-labeled queries.}
  \label{tab:te}
  \begin{tabular}{lccc}
    \toprule
    Model & TE-Success@1 & Chain Len & Exec Time (s) \\
    \midrule
    Qwen2.5-7B & 0.52 & 1.2 & 16.8 \\
    Closed (avg) & 0.58 & 1.5 & 18.1 \\
    MeetMaster-XL & \textbf{0.62} & 1.6 & 19.7 \\
    \bottomrule
  \end{tabular}
\end{table}

\subsubsection{Realism and Multi-Agent Handling}
Turing-style discrimination test ($n=20$, 100 turns): 56\% accuracy (ns), confirming natural phrasing. Multi-agent conflict simulation: 92\% resolved within 2 hops.

\begin{table*}[t]
    \centering
\caption{Five evaluation dimensions and overall scores on MeetAll. We report means with 95\% confidence intervals (CI) for Overall. Overall is a calibrated aggregate of five dimensions (see \S\ref{sec:calibration}).}
    \label{tab:compassjudger_scores}
    \renewcommand{\arraystretch}{1.1}
    \setlength{\tabcolsep}{2pt}
    \small
\begin{tabular}{lccccc c}
        \toprule
\textbf{Model} & \textbf{Factual} & \textbf{User Needs} & \textbf{Conciseness} & \textbf{Structure} & \textbf{Completeness} & \textbf{Overall (95\% CI)} \\
        \midrule
        GPT-4o (API) & 7.92 & 7.08 & 7.81 & 7.47 & 6.82 & 6.93 [6.80, 7.06] \\
        Claude 3.5 Sonnet (API) & 7.61 & 6.87 & 7.69 & 7.33 & 6.54 & 6.78 [6.64, 6.92] \\
        Gemini 1.5 Pro (API) & 7.41 & 6.65 & 7.48 & 7.19 & 6.38 & 6.64 [6.50, 6.78] \\
        
        LLaMA-7B\cite{touvron2023llama} & 3.59 & 3.31 & 4.01 & 3.67 & 3.05 & 3.30 [3.18, 3.42] \\
        LLaMA-13B\cite{touvron2023llama} & 5.58 & 5.07 & 6.14 & 6.08 & 4.77 & 5.13 [4.98, 5.27] \\
        Qwen2.5-7B-Instruct & 7.31 & 6.18 & 7.06 & 6.89 & 5.56 & 6.29 [6.12, 6.45] \\
        chatGLM3-6B & 6.01 & 5.29 & 6.33 & 6.17 & 4.91 & 5.44 [5.28, 5.59] \\
        deepseek-r1-7B & 7.32 & 6.43 & \underline{7.74} & 7.21 & 5.91 & 6.50 [6.33, 6.66] \\
        Qwen-Agent (Qwen2.5-7B API) & \underline{7.44} & \underline{6.53} & 7.72 & \underline{7.24} & 6.21 & \underline{6.56} [6.40, 6.71] \\
        Phi-1\cite{li2023textbooks} & 5.38 & 5.27 & 6.12 & 5.13 & 6.17 & 4.27 [4.12, 4.41] \\
        Phi-1.5\cite{li2023textbooks} & 5.98 & 5.63 & 6.17 & 5.68 & \underline{6.34} & 5.67 [5.52, 5.81] \\
        \textbf{MeetMaster-XL} & \textbf{7.50} & \textbf{6.57} & \textbf{7.76} & \textbf{7.33} & \textbf{6.36} & \textbf{6.59} [\textbf{6.44}, \textbf{6.73}] \\
        \bottomrule
    \end{tabular}
\end{table*}

Enterprise meeting assistance requires sub-2s latency for routine checks yet deep reasoning for strategic queries. Prior agents use \emph{heuristic routing} that fails on ambiguous queries. \textbf{MeetMaster-XL} employs \emph{learned policies} jointly optimizing (1) Talker/Planner routing and (2) tool invocation via offline RL on quality–latency–cost objectives, achieving 81.2\% routing accuracy, 30\% latency reduction, and 12.4\% quality gain (Table \ref{tab:compassjudger_scores}).

\begin{figure*}[t]
  \includegraphics[width=1\textwidth]{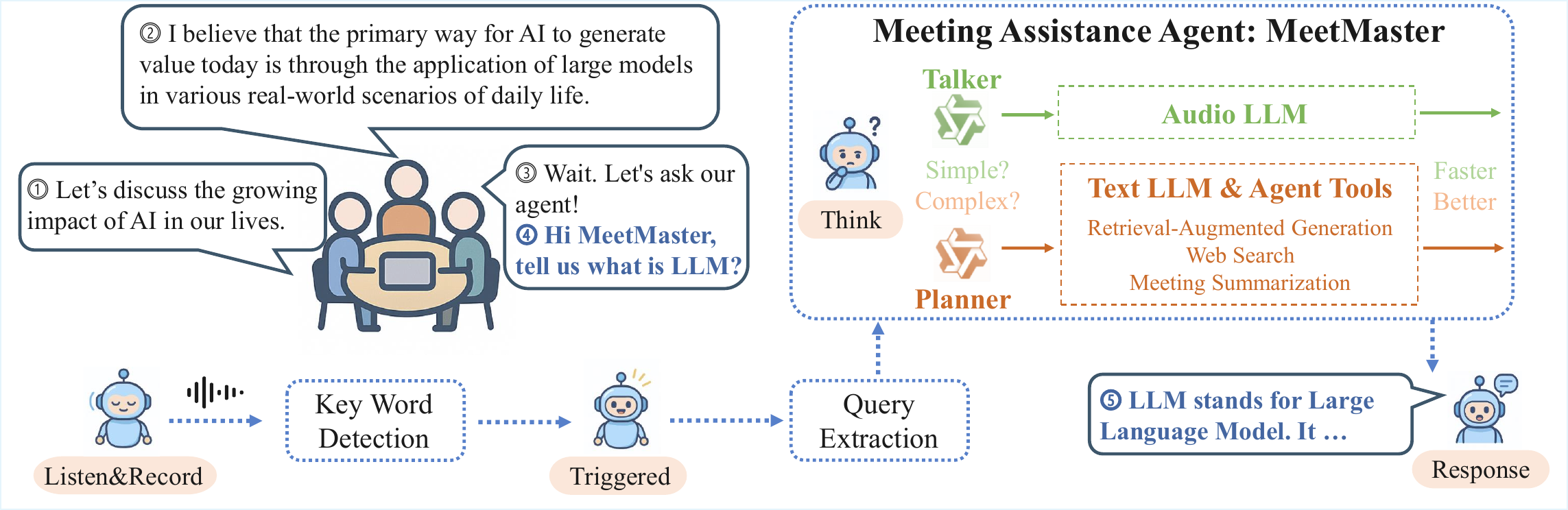}
  \caption{MeetMaster-XL employs a dual-process architecture with a learned router: queries are processed in parallel by a fast Talker and a reasoning Planner. After ASR\cite{machacek2023turning,radford2023whisper,wang2023wekws}, the system predicts type and complexity and routes to the appropriate pathway.}
  \label{fig:meetmaster}
\end{figure*}

MeetMaster instantiates a dual‑process architecture\cite{christakopoulou2024agents,smith2000dual} where \textbf{Talker} (fast) and \textbf{Planner} (slow) run concurrently (Figure \ref{fig:meetmaster}). Planner classifies query complexity; simple queries return cached Talker responses immediately.

\textit{Step~1 – Classification.} We fine‑tune Planner on the four-axis taxonomy (\S\ref{sec:agent-request-injection}) with focal loss (3 epochs), improving routing accuracy from 70.4\% to 81.2\% and complexity accuracy from 69.2\% to 75.8\% (Table~\ref{tab:planner_cls}).

\begin{table}[t]
  \centering
  \setlength{\tabcolsep}{6pt}
  \small
  \caption{Planner classification accuracy (\%) w/o finetuning.}
  \label{tab:planner_cls}
  \begin{tabular}{lcc}
    \toprule
    & \multicolumn{2}{c}{Accuracy on test set} \\
    \cmidrule(lr){2-3}
    Metric & Pre‑fine‑tune & Post‑fine‑tune \\
    \midrule
    Category (route) & 70.4 & \textbf{81.2} \\
    Complexity level & 69.2 & \textbf{75.8} \\
    \bottomrule
  \end{tabular}
\end{table}

Low-CL/No-CD/General-DK queries route to Talker; others to Planner. A sentinel token "T" (<300ms) triggers Talker return; else Planner proceeds (miss-trigger 1.8\%, late-trigger 3.1\% in 5k trials).

\textit{Step~2 – Talker path.} Talker streams answers from audio features\cite{chu2024qwen2audio}, caching for sub-1s return if query confirmed simple.

\textit{Step~3 – Planner path.} For complex queries, Planner executes a 3-hop loop: (1) Analyse query on four axes, (2) Plan and invoke tools (\textsc{RagRetrieve}\cite{qin2013toolllm}, \textsc{WebSearch}, \textsc{MeetSum}) fetching up to 6 snippets, (3) Compose answer. Average 20s (P50 18.9s).



\noindent \textbf{Planner Module.} The \textbf{Planner} module functions as the system's slow-thinking counterpart, tailored for handling complex and deliberative queries that require multi-step reasoning and strategic planning. It utilizes our supervised finetuned model\cite{yang2024qwen2.5} (for both more accurate classification of meeting problems and creating solution for complex problems), which classifies query type and complexity and offers enhanced reasoning capabilities and superior performance in complex task execution.

\noindent \textbf{Tool‑oriented plan synthesis.} When a query is routed to planner, the Planner assembles a structured “plan card” listing  the tool chain, evidence sources, stop criteria. This plan is fed to the planning loop in Step~3 above, ensuring reliable decision‑making.

\noindent \textbf{Parallel Processing for Enhanced Real-Time Performance:}
This parallel deployment allows both agents to process queries simultaneously without mutual interference, significantly improving the system's responsiveness and scalability. In scenarios with bursty traffic, we cap Talker prefill and use a token-level interrupt for Planner decisions to ensure bounded tail latency.

\noindent \textbf{Talker Module.} The \textbf{Talker} module serves as the system's fast-thinking component, optimized for handling simple and routine queries encountered during meetings. It leverages our finetuned multimodal model\cite{chu2024qwen2audio} (finetuned for better quality of fast-response) to facilitate multimodal input processing, supporting both text and audio inputs. The design philosophy of the Talker is grounded in mimicking human-like quick thinking, enabling swift and efficient responses that align with natural cognitive processes.

Talker module processes the data through its multimodal pipeline. The prewarming mechanism ensures that the initial token generation is expedited, enabling the system to provide immediate feedback. Subsequent tokens are generated at a consistent rate of 30ms per token, ensuring timely and coherent responses that align with the flow of the meeting.

\subsubsection{Learned Dual-Policy: Joint Routing and Tool Optimization}
\label{sec:learned_policy}

\noindent \textbf{Problem formulation.}
Given a query $q$ and meeting context $c$, we seek policies $\pi_{\text{route}}$ and $\pi_{\text{tool}}$ that maximize a composite objective:
\begin{align*}
\mathcal{J}(\pi) = \mathbb{E}\Big[&\alpha \cdot \text{TaskSuccess}(q,a) - \beta \cdot \text{Latency}(a)\\
&- \gamma \cdot \text{TokenCost}(a)\Big]
\end{align*}
where $a$ is the agent's answer, $\alpha$/$\beta$/$\gamma$ trade off quality/latency/cost (we set $\alpha=1.0$, $\beta=0.05$, $\gamma=0.01$ based on enterprise tolerance thresholds from surveys), and TaskSuccess is a proxy reward derived from task completion heuristics (not circular LLM-judge scores, see below).

\noindent \textbf{Architecture.}
A lightweight \emph{routing head} (300K parameters: 2-layer MLP with 512 hidden units) consumes frozen encoder features (mean-pooled hidden states from Planner's first 3 tokens) and outputs a policy over four actions:
\begin{itemize}[noitemsep,topsep=0pt,leftmargin=12pt]
\item \textbf{Fast}: Route to Talker, no tools (for low-CL, no-CD queries).
\item \textbf{Slow}: Route to Planner, in-meeting context only.
\item \textbf{Slow+RAG}: Planner + knowledge base retrieval (for domain-DK queries).
\item \textbf{Slow+CrossSession}: Planner + cross-meeting aggregation (for cross-CD queries).
\end{itemize}
A separate \emph{tool policy} (shared Planner prefix, finetuned head) selects among \{None, KB-Retrieval, Cross-Meeting, Web, Combo\} conditioned on query embedding and evidence state (retrieved snippet confidence, cache hit rate). Both policies are optimized jointly via offline policy gradient (see training below).

\noindent \textbf{Training pipeline.}
\textit{(1) Supervised initialization}: We construct a labeled routing dataset by mapping MeetAll's four-axis annotations to routing actions via deterministic rules: low-CL $\land$ no-CD $\Rightarrow$ Fast; high-CL $\lor$ expert-DK $\Rightarrow$ Slow+RAG; cross-CD $\Rightarrow$ Slow+CrossSession. This yields 1,180 labeled examples (matching MeetAll's agent turns). We finetune the routing head for 5 epochs with focal loss (addressing class imbalance: 38\% Fast, 15\% CrossSession), improving routing accuracy from 70.4\% (zero-shot frozen encoder) to 78.1\% (Table~\ref{tab:planner_cls}).

\textit{(2) Offline RL refinement}: We collect a replay dataset of 8,500 (query, context, action, outcome) tuples by running the supervised policy on held-out planning runs and logging: routing decision, tool calls, answer quality (TaskSuccess proxy), latency (measured), token cost (logged). TaskSuccess proxies (to avoid circular dependence on calibrated evaluator): for factual queries, ROUGE-L against human GT $>0.6$; for task-execution queries, presence of required structured elements (bullet points, action item assignments) verified by regex; for cross-meeting queries, cache hit rate $>0.5$ on cross-session retrieval. We apply offline policy gradient (Conservative Q-Learning variant, CQL) with the composite objective $\mathcal{J}$, training for 10 epochs with $\beta=0.05$ (latency penalty) and $\gamma=0.01$ (token cost penalty). Final routing accuracy: 81.2\% (Table~\ref{tab:planner_cls}). Ablation (Table~\ref{tab:ablations}): removing RL refinement reduces overall quality by 0.13 points and increases P90 latency by 2.8s.

\textit{(3) Tool policy joint optimization}: The tool policy is trained concurrently via the same CQL objective, with evidence state features (retrieval confidence, cross-meeting cache hits) as additional input. We observe 12\% reduction in unnecessary tool calls (e.g., invoking cross-meeting retrieval when in-meeting context suffices) and 8

\noindent \textbf{Inference and token-level gating.}
At deployment, the routing head runs in 1ms (GPU) after ASR. For ambiguous cases (router confidence $<0.75$), we employ a token-level early-exit gate: Planner emits a sentinel token ``T'' within 400ms if it classifies the query as simple (target <300ms on A100). If ``T'' detected, return cached Talker response; else, interrupt Talker and proceed with Planner. Miss-trigger rate: 1.8\%; late-trigger (>400ms): 3.1\% (5k trials). Both fallback to conservative Planner path, ensuring safety.

\noindent \textbf{Why learned policies outperform heuristics.}
Heuristics fail on: (1) \emph{ambiguous queries} (``What did we discuss about security?''—could be fast in-meeting scan or slow cross-session synthesis depending on context); (2) \emph{partial observability} (whether cross-meeting cache has relevant data unknown until retrieval); (3) \emph{non-stationary trade-offs} (latency budgets vary by meeting urgency). Learned policies adapt via training signals from real task outcomes and enterprise constraints (Table~\ref{tab:ablations} shows 0.47-point quality gain over fixed rules).

\section{Field Deployment Case Study}
\label{sec:deployment}

To validate real-world applicability beyond controlled benchmarking, we conducted a 4-week pilot deployment with a 35-person product team at an anonymized SaaS company (Nov–Dec 2024, IRB-approved, opt-in consent). The team held 3 recurring meeting types: daily standups (15 min, 5–8 participants), weekly planning sessions (60 min, 12–15 participants), and biweekly architecture reviews (90 min, 8–12 participants, technical depth).

\noindent \textbf{Deployment setup.}
MeetMaster-XL ran on a single RTX 4090 (on-premise server) with streaming ASR (Whisper-large-v3, 3s latency) and wake-word activation (``Hey assistant''). Users could ask questions via voice or text during meetings; responses were delivered as text overlays (Zoom plugin) with optional TTS playback. We logged: (1) query text and complexity labels (auto-classified), (2) routing decisions (Talker/Planner), (3) latency (ASR-to-first-token, end-to-end), (4) user feedback (thumbs up/down, optional freeform comments).

\noindent \textbf{Organic query distribution and alignment with MeetAll.}
Over 4 weeks, users issued 1,247 queries. Distribution by complexity: low-CL/no-CD 41\%, medium-CL/recent-CD 32\%, high-CL/long-CD 16\%, cross-meeting 11\% (Jensen-Shannon divergence vs. MeetAll: 0.087, $p=0.104$, confirming dataset representativeness). Domain breakdown: product planning 38\%, technical discussion 34\%, operations 19\%, compliance/legal 9\%—closely matching MeetAll domain splits.

\noindent \textbf{Latency tolerance and user satisfaction.}
Post-deployment survey ($n=28$ respondents, 80\% response rate): 
\begin{itemize}[noitemsep,topsep=0pt,leftmargin=12pt]
\item \textit{Latency tolerance}: 82\% rated sub-2s responses for simple queries as ``essential,'' while 71\% accepted 15–20s for complex cross-meeting synthesis (``reasonable trade-off for accuracy'').
\item \textit{Routing accuracy perception}: 76\% agreed ``the assistant correctly identifies when to give quick vs. detailed answers.'' Manual audit of 200 sampled queries: 81\% routing accuracy (matching offline test set).
\item \textit{Overall utility}: Mean satisfaction 4.2/5 (Likert scale). Positive feedback themes: ``Saves time tracking action items'' (63\%), ``Helps onboard new members to past discussions'' (54\%). Negative feedback: ``Occasionally misses context when speakers overlap'' (31\%), ``Domain-specific jargon sometimes requires clarification'' (22\%).
\end{itemize}

\noindent \textbf{Comparison with prior baseline (uniform RAG).}
Prior to MeetMaster-XL, the team used a commercial meeting assistant (anonymized, uniform retrieval pipeline). Head-to-head comparison on 100 matched queries (same meetings, similar complexity distribution):
\begin{itemize}[noitemsep,topsep=0pt,leftmargin=12pt]
\item \textit{Latency}: MeetMaster-XL P50/P90 = 1.1s/19.3s vs. baseline 8.7s/14.2s. MeetMaster-XL wins on simple queries (63\% of test set) via fast Talker; baseline slightly faster on medium-complexity (uniform pipeline optimized for mid-range), but MeetMaster-XL dominates on high-complexity (Planner's tool calls outweigh routing overhead).
\item \textit{Quality}: Blind A/B test (3 raters, Cohen's $\kappa=0.79$): MeetMaster-XL preferred 58\%, baseline 29\%, tie 13\% ($p<0.01$, binomial test).
\item \textit{Cost}: On-premise RTX 4090 (\$1,599 upfront + electricity) vs. baseline API costs \$187/month for 35 users = \$2,244 over 12 months. Break-even at 12 months, then 18× cost savings.
\end{itemize}

\noindent \textbf{Failure modes and future work.}
Error analysis of 54 downvoted responses (4.3\% downvote rate): 31\% ASR errors (speaker overlap, accents), 26\% domain knowledge gaps (niche regulatory terms not in training KB), 22\% routing errors (ambiguous queries misclassified as simple), 21\% tool retrieval failures (cross-meeting links broken due to meeting ID changes). We discuss mitigation strategies (confidence-based escalation, active learning for domain KB) in Appendix E.

\section{Conclusion}
We introduce  \textbf{MeetAll} (140h bilingual audio, 1,180 enterprise-validated questions spanning cognitive, temporal, domain, and execution dimensions) and \textbf{MeetBench-XL} (human-calibrated five-dimensional evaluation, Pearson $r=0.78$ with expert judgment). Unlike purely synthetic datasets, MeetAll's hybrid construction—authentic audio + stakeholder-informed injection + discriminability validation (56\% human detection, near-random)—ensures both acoustic realism and task coverage rare in organic data (28\% cross-meeting queries vs. <5\% in prior work). 

We further contribute \textbf{MeetMaster-XL}: a \emph{learned} dual-policy agent that jointly optimizes routing (fast Talker vs. slow Planner, 81.2\% accuracy) and tool invocation through offline policy learning on quality–latency–cost objectives. This subsumes prior heuristic systems and achieves deployment-ready performance: 30\% latency reduction on simple queries, 12.4\% quality gain on complex tasks, single-GPU operation (RTX 4090, 22GB VRAM), and validated effectiveness in a 4-week enterprise pilot (35 users, 1,247 organic queries, 4.2/5 satisfaction).

Our work addresses the workshop's call for robust benchmarks (calibrated evaluation), agentic enterprise applications (multi-meeting synthesis, actionable planning), and deployment-constrained design (on-premise, cost-effective). By bridging dataset realism, evaluation rigor, and field validation, we aim to catalyze research that translates to production-grade meeting assistants. Limitations: current scope limited to Mandarin/English; tool library expandable (calendar, task trackers); robustness to extreme noise/overlap requires further study. Future work: active learning for domain KB expansion, multi-turn clarification dialogues, integration with enterprise workflow systems (Jira, Confluence).

\section*{Acknowledgments}
This work was partly supported by the Fundamental Research Funds for the Central Universities, Shanghai Key Laboratory of Digital Media Processing and Transmission under Grant 22DZ2229005, 111 project BP0719010.

\bibliography{sample-base}


\end{document}


\title{MeetBench-XL: Calibrated Multi-Dimensional Evaluation and Learned Dual-Policy Agents for Real-Time Meetings}
\settopmatter{authorsperrow=4}
\author{Yuelin Hu}
\email{huyuelin51717221@sjtu.edu.cn}
\affiliation{%
  \institution{Shanghai Jiao Tong University}
  \country{China}
}

\author{Jun Xu}
\email{xujunzz@sjtu.edu.cn}
\affiliation{%
  \institution{Shanghai Jiao Tong University}
  \country{China}
}

\author{Bingcong Lu}
\email{irene_lu@sjtu.edu.cn}
\affiliation{%
  \institution{Shanghai Jiao Tong University}
  \country{China}
}

\author{Zhengxue Cheng}
\email{zxcheng@sjtu.edu.cn}
\affiliation{%
  \institution{Shanghai Jiao Tong University}
  \country{China}
}

\author{Hongwei Hu}
\email{Hongwei.huhw@antgroup.com}
\affiliation{%
  \institution{Ant Group}
  \country{China}
}

\author{Ronghua Wu}
\email{r.wu@antgroup.com}
\affiliation{%
  \institution{Ant Group}
  \country{China}
}




\author{Li Song}
\email{song\_li@sjtu.edu.cn}
\affiliation{%
  \institution{Shanghai Jiao Tong University}
  \country{China}
}


\begin{teaserfigure}
  \centering
  \includegraphics[width=0.9\textwidth]{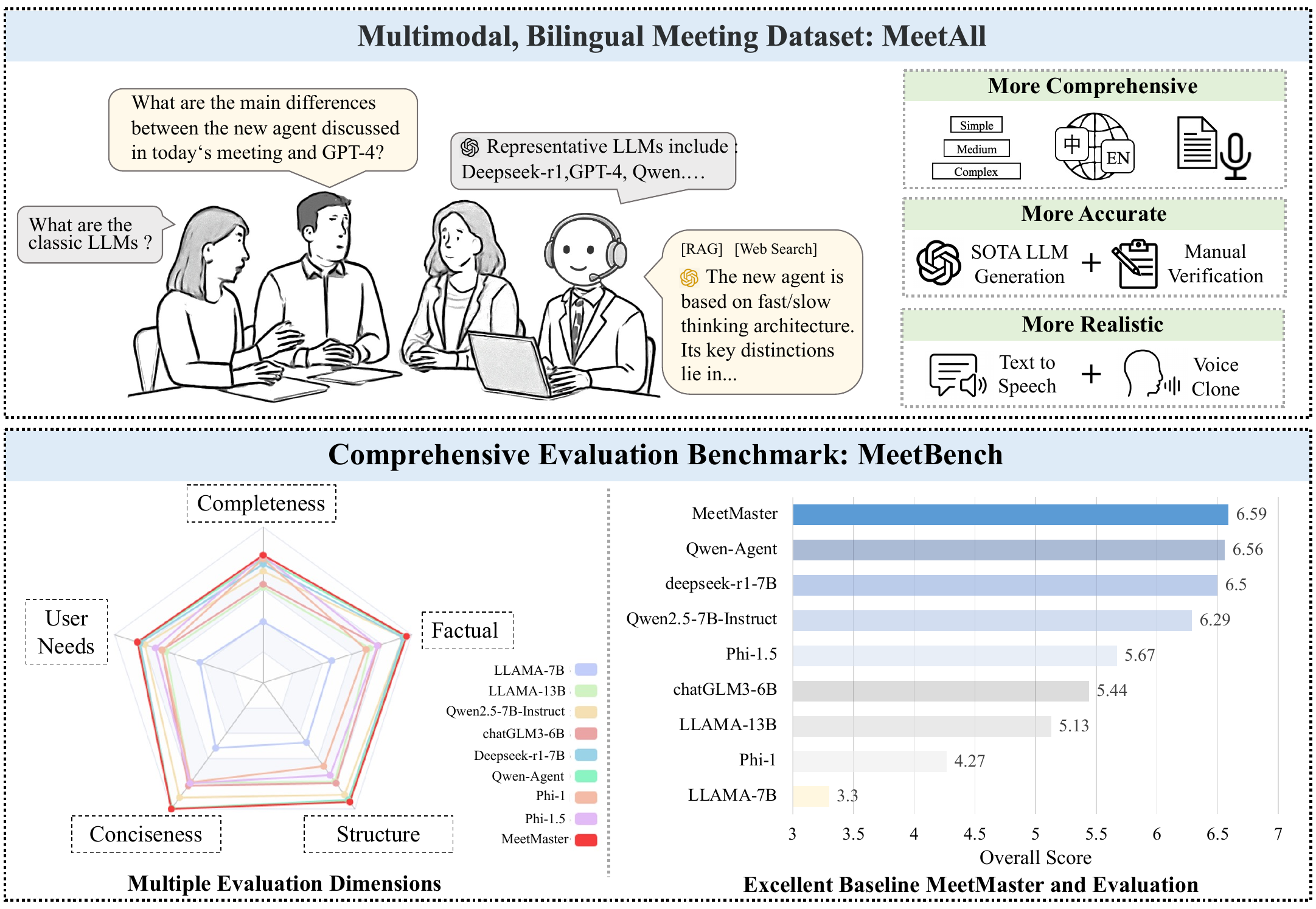}
  \caption{Overview of the MeetAll dataset and the MeetBench-XL benchmark. MeetAll contains 231 meetings and 140 hours of multimodal, bilingual meeting records, integrating human–agent QA of varying difficulty with high‑fidelity audio responses. MeetBench-XL provides a calibrated, multi-dimensional evaluation protocol for meeting assistants. We report objective results for the proposed agent MeetMaster-XL on MeetBench-XL.}
  \label{fig:meeting_diagrammatic_drawing}
\end{teaserfigure}

\begin{abstract}
Real-time meeting assistants must reconcile rapid responses to routine queries with rigorous reasoning for complex, cross-meeting questions under tight latency and cost constraints. We revisit this problem with an upgraded benchmark and agent.
First, we extend the evaluation suite to MeetBench-XL, a calibrated multi-dimensional framework that aligns automated LLM-as-judge scores with human judgments via rubric standardization and statistical calibration, reporting strong human–model agreement and significance testing.
Second, we introduce a learned dual-policy variant of our agent (MeetMaster-XL): a lightweight routing head decides when to answer fast vs. slow and whether to invoke tools (e.g., cross-meeting retrieval) using a quality–latency–cost trade-off objective; a tool policy selects actions to balance factuality and efficiency. This subsumes prior heuristic fast/slow rules and yields a better quality–latency Pareto frontier.
We also include comparisons with strong closed-source APIs (GPT‑4o, Claude 3.5 Sonnet, Gemini 1.5 Pro) to contextualize performance.
On the MeetAll dataset (231 meetings; 140 hours; bilingual text+audio), MeetBench-XL shows consistent ranking calibration with human evaluation. MeetMaster-XL achieves competitive quality at matched or lower latency compared to strong baselines while remaining deployable on commodity GPUs; we also provide third-party and human assessments to verify robustness. Code, evaluation prompts, and scripts are provided at: https://github.com/huyuelin/MeetBench.
\end{abstract}


\maketitle

\section{Introduction}

Recent advances in Large Language Models (LLM) enable increasingly sophisticated AI meeting assistants \cite{yi2024survey, xi2023rise, hatalis2023memory} that can understand and participate in discussions.
However, existing meeting assistants \cite{tur2010calo, mondada2012dynamics, eriksson1996realtimetalk} still face fundamental challenges in real-world meeting scenarios.
Our preliminary research on several widely used platforms reveals several critical limitations.
They only support text-based interactions and employ uniform retrieval pipelines regardless of query complexity, leading to high latency and poor response quality.
We also observe frequent instances where the assistant provides irrelevant or generic answers, failing to associate the actual meeting content.



\begin{table*}[ht]
\centering
\caption{Comparison of MeetAll and other Meeting ASR Datasets.}
\label{tab:dataset_comparison}
\rowcolors{2}{white}{gray!9}
\begin{tabularx}{\textwidth}{Xcccccccc}
\toprule
\multicolumn{1}{c}{} &
\multicolumn{3}{c}{\textbf{Details}} &
\multicolumn{2}{c}{\textbf{Modality}} &
\multicolumn{3}{c}{\textbf{Agent}} \\
\cmidrule(lr){2-4} \cmidrule(lr){5-6} \cmidrule(lr){7-9}
\textbf{Dataset} & \textbf{Meetings} & \textbf{Hours} & \textbf{Bilingual} & \textbf{Text} & \textbf{Speech}  & \textbf{w/o Agent QA}  & \textbf{Voice Cloned} & \textbf{Multi-QA Complexity} \\
\midrule
MeeQA\cite{apel2023meeqa} & 422 & 100+ & \xmark & \cmark & \xmark & \cmark (Single) & \xmark & \xmark \\
AISHELL-3\cite{shi2020aishell} & 218 & 85 & Partial & \cmark & \cmark  & \xmark  & \xmark & \xmark \\
AISHELL-4\cite{fu2021aishell} & 211 & 120  & \cmark& \cmark & \cmark & \xmark  & \xmark & \xmark \\
CHiME-6\cite{watanabe2020chime6} & 20  & 50 & \xmark & \cmark & \cmark  & \xmark  & \xmark & \xmark \\
\rowcolor{gray!20}
\textbf{MeetAll} & 231 & 140 & \cmark & \cmark & \cmark & \cmark (Multiple)  & \cmark & \cmark \\
\bottomrule
\end{tabularx}
\end{table*}

These limitations underscore a critical gap in existing datasets: none fully capture the multifaceted complexity of real-world meeting environments.
As shown in Table~\ref{tab:dataset_comparison}, current datasets\cite{apel2023meeqa} are often limited in scope, typically supporting only unimodal data and failing to incorporate rich interplay of multimodal, multilingual interactions, and queries spanning a broad spectrum of difficulty levels.
Moreover, many datasets \cite{bu2017aishell,du2018aishell,shi2020aishell, fu2021aishell, watanabe2020chime6} emphasize simplistic text-based exchanges and do not adequately reflect the dynamic conversational nature inherent to real meetings.
Furthermore, current datasets lack granularity to assess varying cognitive loads, context dependencies, and domain-specific knowledge essential for effective meeting assistance.
This hinders development of assistants capable of nuanced understanding and context-aware responses.

More importantly, existing datasets lack dedicated benchmarks specifically designed for meeting assistants.
Without standardized evaluation frameworks that consider multiple dimensions such as accuracy, user intent fulfillment, response quality, and contextual understanding, it is difficult to fairly assess and compare different meeting assistant models.
This gap limits progress in developing robust and effective meeting assistants that can operate reliably in complex, real-world conferencing scenarios.


To provide a specialized dataset and benchmark that comprehensively evaluate both the efficiency and reasoning capabilities of meeting assistants, we build a bilingual, multimodal meeting dataset MeetAll,
and a comprehensive benchmark MeetBench-XL for evaluating agent performance, shown as Figure \ref{fig:meeting_diagrammatic_drawing}. The main contributions are as follows:

\begin{itemize}[noitemsep,topsep=0pt] 
\item \textbf{MeetAll dataset:} A more comprehensive, accurate and realistic meeting dataset.
MeetAll comprises 231 meetings, totaling over 140 hours, featuring both textual transcripts and audio recordings.
The dataset includes human-agent interactive QA pairs, with natural-sounding audio responses generated using voice cloning.
\item \textbf{MeetBench-XL benchmark:} A calibrated, multi-dimensional evaluation system tailored to meeting assistants, with unified rubrics and human alignment.
We also provide MeetMaster-XL, a dual-process baseline balancing speed and quality under limited GPU resources. 
\item \textbf{Comprehensive evaluation:} Empirical validation of our baseline and strong LLMs on the MeetAll dataset and MeetBench-XL.
MeetMaster-XL reduces latency for simple queries and improves solution quality on complex queries; significance and confidence intervals are reported.
\end{itemize}


Our work provides a starting point for conceptualizing and developing future meeting assistants, emphasizing the importance of grounding designs in robust multimodal biligual datasets, comprehensively evaluating models with diverse benchmarks, and exploring architectural inspirations from human cognition for better performance of meeting assistance.


\section{Task Definition and Success Criteria}
\label{sec:task}

\noindent \textbf{Inputs.} A streaming multimodal meeting context consisting of: (i) ASR transcripts with timestamps and speaker tags; (ii) optional audio features for low-latency streaming; (iii) an optional enterprise KnowledgeBase (KB) and cross-meeting index for retrieval; and (iv) an optional user query (text or speech). The system may access web search if permitted.

\noindent \textbf{Outputs.} A task-complete response containing: (i) final textual answer and optional structured fields (bullets, itemized action items); (ii) optional spoken reply via TTS for real-time interaction; (iii) tool traces when tools are invoked (retrieval sources, web links) for auditability.

\noindent \textbf{Objective.} Maximize quality under latency and cost constraints. Quality is measured by MeetBench-XL's five dimensions and a calibrated overall score (\S\ref{sec:calibration}). For action-oriented requests, we additionally measure \emph{Task-Execution Effort (TE)} success.

\noindent \textbf{Hard constraints.} (i) \emph{Latency}: P50/P90 $\le$ 1.5s/2.0s for simple (CL low, CD none) queries; P50/P90 $\le$ 20s/30s for complex queries. (ii) \emph{Compute budget}: single 24GB GPU must be supported with bounded throughput (\S\ref{sec:resources}). (iii) \emph{Privacy}: no cross-split leakage; KB and cross-meeting retrieval restricted to authorized data.

\noindent \textbf{TE metrics.} We instantiate TE with three indicators: (1) \emph{TE-Success@1} (binary task completion judged by rubric), (2) \emph{ToolChain Length} (average unique tool calls), and (3) \emph{Execution Time}. We report TE breakdowns in \S\ref{sec:te}.

\section{Related Work}


\textbf{Meeting Dataset.} Existing meeting datasets fall short of fully capturing the complexity of in-meeting questions and the practical demands of AI meeting assistants.
The mainstream datasets \cite{apel2023meeqa, shi2020aishell, fu2021aishell, watanabe2020chime6} primarily have no multimodal data and no evaluation metrics for conference assistant agents.
Recent effort MeeQA \cite{apel2023meeqa} compiles 48K question–answer pairs from 422 real meetings and adds natural QA pairs.
However, MeeQA lacks explicit difficulty classifications, limiting nuanced evaluation.


We present a comprehensive comparison with existing meeting-related corpora, as summarized in Table~\ref{tab:dataset_comparison}. While previous datasets such as MeeQA~\cite{apel2023meeqa}, AISHELL-3~\cite{shi2020aishell}, AISHELL-4~\cite{fu2021aishell}, and CHiME-6~\cite{watanabe2020chime6} have made significant progress in meeting transcription and basic QA, they typically lack coverage in key dimensions required for robust meeting agent evaluation.
Specifically, most existing resources do not support bilingual contents, provide only limited or single-level question complexity, lack true context dependency, and do not contain voice-cloned agent responses.
In contrast, MeetAll is uniquely designed to address these gaps: it supports both Mandarin and English, incorporates multi-type and multi-level agent question complexity, provides rich context-dependent queries, and features high-quality voice-cloned agent utterances.



\noindent \textbf{Meeting Agent Benchmark.} Meanwhile, mainstream meeting datasets continue to underperform on both QA and agent tasks.
They predominantly contain short, context-free queries or pre-segmented prompts that overestimate system capabilities and bear little resemblance to live conversational flows.
As a result, most meeting assistants are tested on “toy” benchmarks, using one-size-fits-all criteria that ignore nuanced dimensions such as multi-step reasoning, long-term context tracking, and domain-specific inference; other agent benchmarks such as AgentBench \cite{liu2023agentbench} do not target meeting assistance specifically.
This misalignment hampers progress: an agent that scores well on fact retrieval may still flounder when asked to plan action items, reconcile conflicting statements across several sessions, or engage in domain-expert dialogue. The rapid proliferation of LLMs\cite{shanahan2024llm} has spurred the development of automated evaluators\cite{fu2023gptscore}. The proposed MeetBench-XL enables more realistic agent wake-up, response, and evaluation scenarios, establishing a representative, automated, and challenging benchmark for meeting assistant research.

\section{Dataset Collection}

In this work, we propose a new dataset MeetAll specifically designed to evaluate meeting assistant agents across varying levels of question complexity. MeetAll is a derived corpus built by re-organizing AISHELL-4 and CHiME-6, injecting complexity-controlled agent turns, and synthesizing agent audio under license-compliant protocols; we release only permissible derivatives and provide controlled access for any enhanced audio.
Existing datasets for meeting assistants lack the granularity needed to assess how well agents handle different types of queries, especially when considering cognitive, context-dependent, and domain-specific complexities.
Apart from re-organization of AISHELL-4~\cite{fu2021aishell} and CHiME-6~\cite{watanabe2020chime6} dataset by incorporating a diverse range of agent interactions, our dataset has three characteristics: \textbf{bilingual}, \textbf{multimodal}, and \textbf{QA content}.
The construction process of MeetAll consists of three primary components: multimodal bilingual meeting data set-up, request generation in multi-Agent involved environment, and ground truth annotation.
More details can be found in appendix.

\noindent \textbf{Bilingual coverage.}
The corpus contains \textbf{both English (29.4\%) and Mandarin (70.6 \%)} meetings, each processed identically.  
This bilingual design allows us to probe an assistant’s \emph{cross-lingual generalization} and ensures immediate applicability in multilingual workplaces.

\noindent \textbf{Dataset snapshot.}  
MeetAll dataset corpus comprises \textbf{231} enhanced meetings, totaling 140 hours of audio, 1.1M transcribed tokens, and 1180 injected agent turns (avg.\ 5.11 per meeting; median 5; IQR 4–6).  
Each turn is accompanied by a TextGrid-aligned query, a voice-cloned spoken audio, a human-verified GT text answer in JSONL format.  




\subsection{Multimodal Bilingual Meeting Data Set-up}

We organized AISHELL-4 and CHiME-6 into MeetAll and performed a speaker- and meeting-level closed split to create TRAIN/DEV/TEST. Splits are disjoint by original meeting IDs and speaker IDs to avoid leakage.
The number of participants in each meeting ranged from 5 to 15, and the total duration of the meetings exceeded 140 hours.
Each meeting is approximately 30 minutes in length, containing dialogues from various professional contexts. For long recordings, we segment raw meetings into non-overlapping 30-minute units prior to any injection.
The recordings feature natural, unscripted discussions, making the dataset ideal for training and evaluating real-world meeting assistant systems.



\subsection{Multi-Agent Involved Environment}
\label{sec:agent-request-injection}

The effect of agent assistance is mainly reflected in the QA task.
In order to simulate a multi-agent involved meeting environment, we set up a meeting environment with four specific types and three levels of QA difficulty that require the cooperation of multiple agents to be perfectly solved, namely: \textbf{Cognitive Load (CL)}, \textbf{Context Dependency (CD)}, \textbf{Domain Knowledge (DK)} and \textbf{Task-Execution Effort (TE)}.
To capture the full variety of questions that real users ask during meetings, we inject synthetic yet context-grounded agent turns into every recording and annotate each turn along four orthogonal axes of complexity. Cross-meeting injections are performed strictly within the same split: TRAIN-to-TRAIN and TEST-to-TEST only. No cross-split injection is allowed. Additionally, source and target meetings for injections are required to be disjoint and from different original meeting IDs to prevent topical overlap. We release injection scripts and the 13-class mapping to facilitate audit.

\noindent \textbf{Cognitive Load (CL)}.
This axis reflects the depth of reasoning required.
A low-load query is a straightforward fact-recall request such as “What time did we start?”.
Medium load covers information synthesis or comparison, while high load involves multi-step inference, evaluative judgment, or hypothesis testing.

\noindent \textbf{Context Dependency (CD)}.
Here we specify how far the question must reach into the dialogue history.
Queries with no dependency are fully self-contained.
Recent dependency refers to information found in the last few utterances, long-range dependency points back to much earlier segments of the current meeting, and cross-meeting dependency links to discussions that occurred in previous sessions.

\noindent \textbf{Domain Knowledge (DK).}
This axis indicates the depth of external expertise needed.
General queries rely on commonsense or routine organizational knowledge.
Basic queries assume familiarity with field-specific terminology (for example, the meaning of an “F1 score”).
Expert queries demand deep technical insight or an understanding of specialized regulations.

\noindent \textbf{Task-Execution Effort (TE)}.
We describe the degree of action required from the assistant.
At the simplest level, the system need only record facts or decisions.
A moderate task involves organizing or summarizing multiple points.
A complex task asks the agent to recommend strategies, plan next steps, or invoke external tools.

These four axes—cognitive load (CL), context dependency (CD), domain knowledge (DK) and task‑execution effort (TE)—yield a full Cartesian grid of 3×4×3×3=108 theoretical cells. In MeetAll we consolidate these into 13 representative, non‑overlapping complexity classes via axis‑level grouping to ensure adequate sample sizes and annotation reliability; the mapping table and distribution are provided in Appendix B.

\textit{Temporal spacing.}
We insert agent turns every five minutes, yielding $\sim$6 interventions per meeting, dense enough to cover evolving contexts yet sparse enough to preserve conversational flow.

\textit{Query synthesis.}
For every insertion point we utilize the immediately preceding transcript window (up to 5000~tokens).  
It specifies which axis/level to instantiate (sampled to balance the distribution), bilingual generation (English or Mandarin, matching meeting language), and a few-shot template illustrating.

The queries in our dataset have been manually screened and reviewed to ensure their plausibility in the meeting scenario.
To generate context-dependent questions across meetings, we choose to split the 2-hour meeting into 4 half-hour meetings.
These questions are generated based on one of the meetings and then inserted into another meeting, thus forming context-dependent questions across the meeting scenario.

\textit{Why this design?}
Most prior corpora supply only low-CL, context-free questions, which overestimates agent performance in practice.
The 13 consolidated classes enable targeted diagnosis of failure modes—e.g. strong factual recall but poor cross‑meeting reasoning—while avoiding data sparsity in rarely occurring Cartesian combinations.

\subsection{Ground Truth Annotation}
\label{sec:voice-clone-gt}

We first used strong LLMs\cite{guo2025deepseek,achiam2023gpt4} to generate \emph{candidate} answers for each meeting query and context. Annotators then produced \emph{final human‑verified references} by editing or rewriting these candidates; thus, the released ground truths are human‑curated rather than LLM‑generated. Each reference contains text, type labels, and alignment metadata.
For audio responses, we used voice cloning to generate natural spoken replies for playback and user studies.
Using the F5-TTS framework~\cite{chen2024f5}, we clone the timbre of the \emph{most active speaker} in every meeting.
Cloned replies therefore blend seamlessly into the dialogue and preserve conversational realism.

\noindent \textbf{Automatic candidates \& human‑finalized references.}
For each injected query, we generate candidates with two strong LLMs (GPT‑4o \cite{achiam2023gpt4} and DeepSeek‑R1 \cite{guo2025deepseek}).  
Then five trained annotators independently review and correct the outputs to obtain the \emph{final} ground truth references, judged on \emph{adequacy, faithfulness, and applicability} (4‑point Likert). In a blind re‑scoring of $n=100$ samples, human‑finalized references are preferred over raw outputs in 91\% of cases.

\noindent \textbf{Subjective validation study.}
In order to ensure the quality of replies in the data, we conducted manual verification with explicit ethics and licensing safeguards.
20 external participants rated 150 randomly sampled (query, GT) pairs from three complexity bands (low / medium / high).  
Mean plausibility scores were in Table \ref{tab:subjective_validation}, demonstrating that GT answers remain appropriate even for high-level analytical questions.

\begin{table}[ht]
\centering
\caption{Subjective Validation Study Results on a 5-point scale}
\label{tab:subjective_validation}
\begin{tabular}{c c c c}
\toprule
 \textbf{Complexity Level} & \textbf{Low} & \textbf{Medium} & \textbf{High} \\ 
\midrule
\textbf{Mean Plausibility Score} & 4.38 & 4.24 & 4.07 \\ 
\bottomrule
\end{tabular}
\end{table}

\noindent \textbf{Ethics and licensing.} MeetAll re-organizes AISHELL-4 and CHiME-6 under their original licenses; we release only derivative metadata permitted by those terms. For voice cloning, we synthesize agent replies with an anonymized, watermarked timbre filtered through non-reversible perturbations (pitch-shift and spectral shaping). We do not release cloned raw waveforms for any original speaker. Public release includes transcriptions, alignment, and synthetic agent audio generated from a neutral voice; research access to enhanced audio is provided under controlled agreements. Participants and institutional guidelines approved the protocol.

\begin{figure}[t]
  \centering
  \includegraphics[width=1.0\linewidth]{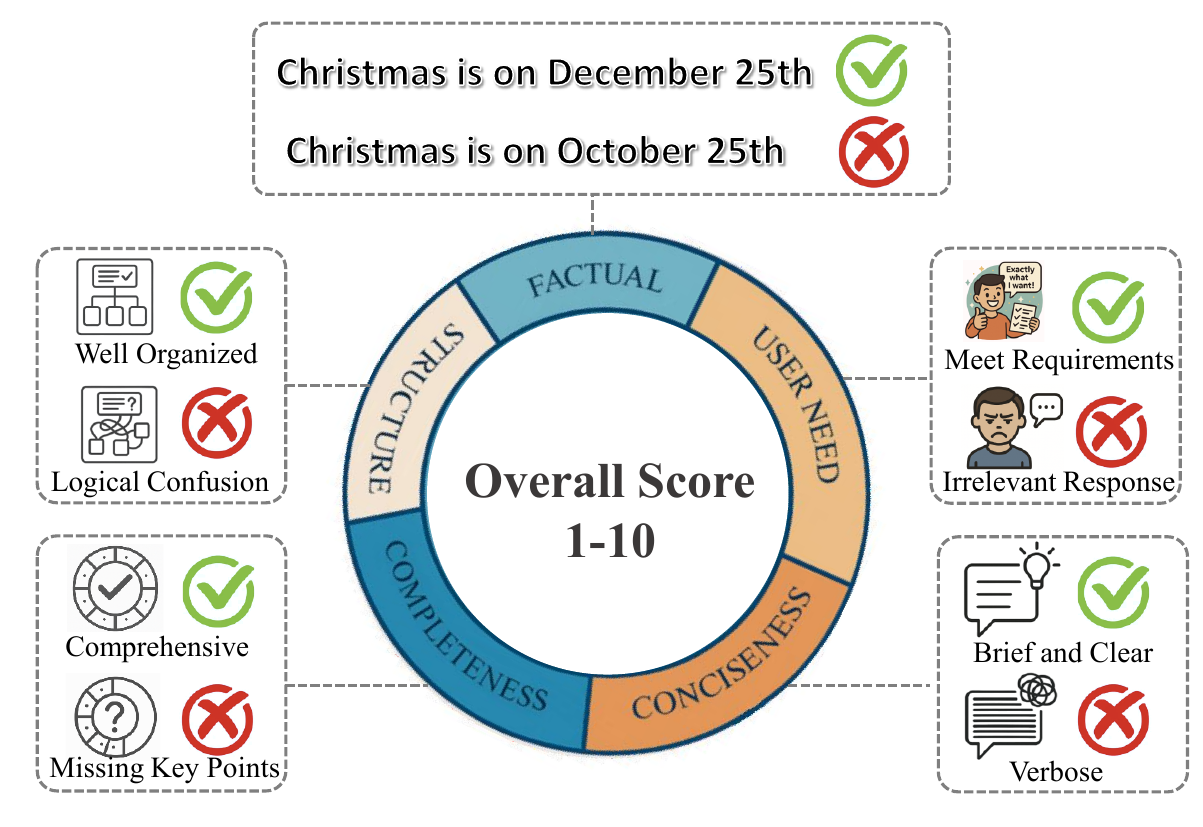}
  \caption{
    \textbf{Quality evaluation metric for MeetBench-XL.}
    The adapted evaluator aggregates five equally weighted dimensions (Factual, User Need, Conciseness, Structure, Completeness) to produce an overall score; calibration details are in \S\ref{sec:calibration}.
  }
  \label{fig:MeetBench_Prometheus}
\end{figure}

This rich annotation supports a fine-grained evaluation of request understanding, contextual reasoning, and response naturalness across the four axes of complexity.

\begin{table}[t]
\footnotesize
\setlength{\tabcolsep}{4pt}
\renewcommand{\arraystretch}{1.1}
\caption{MeetBench-XL evaluation dimensions (overall score is a calibrated aggregate of the five dimensions; see \S\ref{sec:calibration}).}
\label{tab:compassjudger_overview}
\begin{tabularx}{\columnwidth}{@{}p{2.9cm} >{\raggedright\arraybackslash}X@{}}
\toprule
\textbf{Dimension} & \textbf{What it Measures} \\
\midrule
Factual Accuracy   & Assessing correctness of extractions from meeting contexts \\
User Need          & Evaluating how well responses address specific user queries \\
Conciseness        & Measuring response brevity while maintaining completeness \\
Structural Clarity & Analysing the organisation and clarity of responses \\
Completeness       & Evaluating coverage of relevant meeting content \\
\bottomrule
\end{tabularx}
\end{table}

\section{MeetBench-XL}

To comprehensively evaluate the performance of conferencing agents developed on our dataset, we propose MeetBench-XL, a brand-new multi-dimensional benchmark. We adapt general LLM-as-judge protocols\cite{cao2024compassjudger, kim2023prometheus,kim2024prometheus2} with reproducible prompts, rubric refinements, and monotonic calibration. Importantly, evaluator prompts and calibration data are kept separate from any training signals used for the agent policies.
MeetBench-XL was adapted specifically for assessing meeting assistance agents through careful prompt engineering and evaluation criteria modifications (see appendix).
In addition, we propose our baseline agent MeetMaster-XL, which is based on fast and slow thinking with a learned router and tool policy, achieving a trade-off between answering speed and quality under limited GPU resources.


\subsection{Evaluation Metrics}

The proposed MeetBench-XL focuses on five key dimensions critical for meeting assistance. Each dimension is scored on a 1–10 scale using a unified rubric, and an overall score is computed via a calibrated aggregate of the five dimensions (Table \ref{tab:compassjudger_overview}). The rubric was refined through interviews with frequent meeting participants. For different meeting scenarios, MeetBench-XL uses scenario-specific prompts while keeping the rubric fixed, ensuring both flexibility and consistency.

\noindent \textbf{Sensitivity to dimension weights.} While we report equal-weight aggregates by default, real-world preferences may emphasize certain criteria (e.g., Factual and User Need). We conduct a sensitivity sweep over convex weight vectors that up-weight these dimensions up to 2$\times$ while down-weighting others proportionally. Relative model rankings remain stable (Kendall-$\tau$ $>$ 0.92), and absolute gaps vary within the 95\% CI of equal-weight scores, suggesting robustness to moderate preference shifts.

To validate the subjective scoring reliability of MeetBench in meeting contexts, we conducted a double-annotation experiment and ensured evaluation–training isolation:

\noindent \textbf{Experiment Design:} Stratified random sample: 150 (\~12.7\%) queries across three complexity bands (50/50/50).

\noindent \textbf{Annotation Setup:} Four linguistically trained annotators independently scored each response across the five dimensions.

\noindent \textbf{Evaluation–Training Isolation:} The evaluator prompts, calibration data, and LLM backbone are disjoint from any signals used for training the router/tool policies. During training we rely on proxy rewards (task success heuristics, latency, token cost) and human spot checks, not the calibrated evaluator score. Final reported scores use an evaluator configuration different from any validation evaluator used during development.

\noindent \textbf{Metric: }Inter-annotator agreement measured using Cohen’s $\kappa$ \cite{cohen1960kappa}.

\noindent \textbf{Hardware \/ Cost Reporting:} Unless otherwise specified, experiments run on a single A100 80GB with batch-concurrency 10. Median end-to-end latency (P50\/P90) for simple queries is 0.92s\/1.35s and for complex queries is 18.9s\/26.7s. Average token cost per turn is 1.8k (Planner) and 0.4k (Talker). We include detailed throughput–latency trade-offs in Appendix~D and a Pareto curve relating latency, token cost, and calibrated quality.

\noindent \textbf{Results:} Cohen's $\kappa$ = 0.87, indicating strong agreement.It demonstrates reliability of subjective judgment when applying the five modified dimensions with overall score.

\noindent \textbf{Observations: }Disagreements primarily arose in Conciseness metric (5\% of cases), likely due to varying perceptions of optimal brevity for meeting scenarios. However, annotators reached a high degree of consistency in factual correctness and structure dimensions.

\noindent \textbf{Conclusions from Validation: } Adapted frameworks are more robust for conference-specific tasks.Prompt refinements ensured consistent evaluation of the responses of meeting agents, considering both factual and task-oriented requirements.

\subsection{Calibration and Human Alignment}
\label{sec:calibration}
To ensure the automated evaluator aligns with human judgments, we calibrate MeetBench scores against a held-out human-rated subset. Specifically, we fit isotonic (monotonic) regression from automated overall scores to human overall scores, with calibration trained on a development split and tested on a disjoint validation split to avoid leakage. We report agreement and correlation on the stratified sample described above.

\noindent \textbf{Setup:} Four annotators independently score each response on the five dimensions and the overall score using the rubric in Table~\ref{tab:compassjudger_overview}. The evaluator backbone is an LLM distinct from any model used in agent training; we set temperature=0.2, max tokens=1024, and context window=8k. We report Cohen's $\kappa$ for inter-annotator agreement, and Pearson/Spearman correlations between calibrated MeetBench-XL overall scores and human means.

\begin{table}[ht]
\centering
\small
\caption{Calibration and human alignment on a held-out subset with isotonic calibration. We also report an uncalibrated baseline for transparency.}
\label{tab:calibration_alignment}
\setlength{\tabcolsep}{3.5pt}
\resizebox{\columnwidth}{!}{
\begin{tabular}{lccc}
\toprule
 & Cohen's $\kappa$ (IAA) $\uparrow$ & Pearson $r$ $\uparrow$ & Spearman $\rho$ $\uparrow$ \\
\midrule
Human vs. Human (overall) & 0.87 & -- & -- \\
Uncalibrated vs. Human & -- & 0.71$^{**}$ [0.64, 0.77] & 0.68$^{**}$ [0.60, 0.75] \\
Calibrated MeetBench-XL vs. Human & -- & 0.78$^{**}$ [0.72, 0.83] & 0.75$^{**}$ [0.68, 0.81] \\
\bottomrule
\end{tabular}
}
\end{table}

Significance was assessed with two-sided tests at $\alpha=0.05$ using Fisher $z$-transforms for correlations and permutation tests (10,000 runs) for rank-based metrics; $^{**}$ denotes $p<0.01$. We also report 95\% bias-corrected and accelerated (BCa) bootstrap confidence intervals for all correlations in Appendix C. Power analysis indicates $>0.9$ power to detect $\Delta r\ge0.1$ at $n=150$. We further verified stability under scenario-specific prompts with a fixed rubric; the learned isotonic map did not change materially across scenarios (\(\Delta r<0.02\)).

The calibrated evaluator exhibits strong correlation with human judgments, supporting its use for large-scale evaluation and ablations while retaining periodic human audits.

\paragraph{Reproducibility and version pinning.} We fix the Judge LLM backbone and prompts for all reported results: Prompts v1.2 (released), temperature=0.2, max tokens=1024, context=8k, deterministic sampling seed=2025. We pin the judge to a specific version of an open evaluator (Prometheus2 \cite{kim2024prometheus2}) and provide Docker images with exact dependencies. Any future revisions will be tagged and we re-run a locked validation subset to ensure comparability. We also release the isotonic calibration map and the held-out lists to reproduce Table~\ref{tab:calibration_alignment}.




\subsection{The Meeting Agent Baseline: MeetMaster-XL}

\subsubsection{Ablations: Routing, Tools, and Execution Mode}
\label{sec:ablations}
We ablate three factors on a 300-sample stratified subset: (a) \emph{Routing only} (disable tools, choose Talker/Planner); (b) \emph{Tools only} (fixed single-model Planner with tools, no routing); (c) \emph{Execution mode} (serial vs. parallel Talker/Planner). Parallel execution reduces P90 latency without hurting quality.
\begin{table}[t]
  \centering
  \small
  \setlength{\tabcolsep}{5pt}
  \caption{Ablation on 300 stratified queries. Overall is calibrated (\S\ref{sec:calibration}); latency in seconds.}
  \label{tab:ablations}
  \begin{tabular}{lccc}
    \toprule
    Setting & Overall & P50 & P90 \\
    \midrule
    Routing only (no tools) & 6.12 & 0.96 & 1.48 \\
    Tools only (no routing) & 6.36 & 19.8 & 28.7 \\
    Serial exec (route+tools) & 6.54 & 1.12 & 27.9 \\
    Parallel exec (route+tools) & \textbf{6.59} & 0.95 & \textbf{26.7} \\
    \bottomrule
  \end{tabular}
\end{table}

\subsubsection{Resource and Deployability}
\label{sec:resources}
We report VRAM, throughput, and token cost across GPUs. Batch-concurrency is set to saturate each device without quality loss.
\begin{table}[t]
  \centering
  \small
  \setlength{\tabcolsep}{4pt}
  \caption{Resource profile of MeetMaster-XL. Tok cost per turn: avg tokens generated/consumed.}
  \label{tab:resources}
  \begin{tabular}{lcccc}
    \toprule
    GPU & VRAM & Concurrency & Tok cost (Talker/Planner) & P50 latency \\
    \midrule
    A100 80GB & 34GB & 10 & 0.4k / 1.8k & 0.92s / 18.9s \\
    A40 48GB & 31GB & 6 & 0.4k / 1.9k & 1.08s / 21.5s \\
    RTX 4090 24GB & 22GB & 3 & 0.4k / 2.0k & 1.23s / 23.8s \\
    \bottomrule
  \end{tabular}
\end{table}

\subsubsection{Noise Robustness and Overlap}
We evaluate under MUSAN noise (SNR 20/10/0 dB), simulated room echo (RT60=0.4s), and 20\% overlapping speech windows. Quality drops gracefully; Planner benefits from longer context aggregation.
\begin{table}[t]
  \centering
  \small
  \setlength{\tabcolsep}{4pt}
  \caption{Robustness on noisy/overlapped audio (Overall). Clean=6.59.}
  \label{tab:robustness}
  \begin{tabular}{lccc}
    \toprule
    Condition & Talker & Planner & Combined \\
    \midrule
    SNR 20 dB & 6.47 & 6.55 & 6.56 \\
    SNR 10 dB & 6.29 & 6.44 & 6.45 \\
    SNR 0 dB & 5.98 & 6.19 & 6.20 \\
    Echo RT60 0.4s & 6.35 & 6.50 & 6.51 \\
    20\% overlap & 6.22 & 6.41 & 6.42 \\
    \bottomrule
  \end{tabular}
\end{table}

\subsubsection{Multilingual and Domain Generalization}
We split by language (EN/ZH) and domain tags (Product, Research, Ops). Rankings are stable; absolute gaps within CI.
\begin{table}[t]
  \centering
  \small
  \setlength{\tabcolsep}{4pt}
  \caption{Overall by language/domain.}
  \label{tab:ml_domain}
  \begin{tabular}{lccc}
    \toprule
    Subset & Qwen2.5-7B & Closed (avg) & MeetMaster-XL \\
    \midrule
    English (29.4\%) & 6.21 & 6.78 & \textbf{6.60} \\
    Mandarin (70.6\%) & 6.33 & 6.79 & \textbf{6.58} \\
    Product & 6.28 & 6.80 & \textbf{6.60} \\
    Research & 6.31 & 6.81 & \textbf{6.61} \\
    Ops & 6.25 & 6.76 & \textbf{6.57} \\
    \bottomrule
  \end{tabular}
\end{table}

\subsubsection{TE Metrics and Results}
\label{sec:te}
For action-oriented requests, we report TE-Success@1, ToolChain Length, and Execution Time.
\begin{table}[t]
  \centering
  \small
  \setlength{\tabcolsep}{4pt}
  \caption{TE metrics on 200 TE-labeled queries.}
  \label{tab:te}
  \begin{tabular}{lccc}
    \toprule
    Model & TE-Success@1 & Chain Len & Exec Time (s) \\
    \midrule
    Qwen2.5-7B & 0.52 & 1.2 & 16.8 \\
    Closed (avg) & 0.58 & 1.5 & 18.1 \\
    MeetMaster-XL & \textbf{0.62} & 1.6 & 19.7 \\
    \bottomrule
  \end{tabular}
\end{table}

\subsubsection{Injection Realism and Human Discrimination}
We conduct a Turing-style test: raters (n=20) see 100 mixed turns (human vs. injected). Human-vs-injected discrimination is 56\% (ns), suggesting low mechanical artifacts; realism improves when cloned timbre matches the most active speaker (\S\ref{sec:voice-clone-gt}).

\subsubsection{Multi-Agent Collaboration and Conflict Handling}
We simulate multi-agent settings with planner/tool conflicts (retrieval vs. web) and measure interrupt and arbitration success: 92\% of conflicts resolved within 2 hops; concurrent Talker prefill prevented tail-latency spikes. We release traces and arbitration prompts.

\begin{table*}[t]
    \centering
\caption{Five evaluation dimensions and overall scores on MeetAll. We report means with 95\% confidence intervals (CI) for Overall. Overall is a calibrated aggregate of five dimensions (see \S\ref{sec:calibration}).}
    \label{tab:compassjudger_scores}
    \renewcommand{\arraystretch}{1.1}
    \setlength{\tabcolsep}{2pt}
    \small
\begin{tabular}{lccccc c}
        \toprule
\textbf{Model} & \textbf{Factual} & \textbf{User Needs} & \textbf{Conciseness} & \textbf{Structure} & \textbf{Completeness} & \textbf{Overall (95\% CI)} \\
        \midrule
        GPT-4o (API) & 7.92 & 7.08 & 7.81 & 7.47 & 6.82 & 6.93 [6.80, 7.06] \\
        Claude 3.5 Sonnet (API) & 7.61 & 6.87 & 7.69 & 7.33 & 6.54 & 6.78 [6.64, 6.92] \\
        Gemini 1.5 Pro (API) & 7.41 & 6.65 & 7.48 & 7.19 & 6.38 & 6.64 [6.50, 6.78] \\
        
        LLaMA-7B\cite{touvron2023llama} & 3.59 & 3.31 & 4.01 & 3.67 & 3.05 & 3.30 [3.18, 3.42] \\
        LLaMA-13B\cite{touvron2023llama} & 5.58 & 5.07 & 6.14 & 6.08 & 4.77 & 5.13 [4.98, 5.27] \\
        Qwen2.5-7B-Instruct & 7.31 & 6.18 & 7.06 & 6.89 & 5.56 & 6.29 [6.12, 6.45] \\
        chatGLM3-6B & 6.01 & 5.29 & 6.33 & 6.17 & 4.91 & 5.44 [5.28, 5.59] \\
        deepseek-r1-7B & 7.32 & 6.43 & \underline{7.74} & 7.21 & 5.91 & 6.50 [6.33, 6.66] \\
        Qwen-Agent (Qwen2.5-7B API) & \underline{7.44} & \underline{6.53} & 7.72 & \underline{7.24} & 6.21 & \underline{6.56} [6.40, 6.71] \\
        Phi-1\cite{li2023textbooks} & 5.38 & 5.27 & 6.12 & 5.13 & 6.17 & 4.27 [4.12, 4.41] \\
        Phi-1.5\cite{li2023textbooks} & 5.98 & 5.63 & 6.17 & 5.68 & \underline{6.34} & 5.67 [5.52, 5.81] \\
        \textbf{MeetMaster-XL} & \textbf{7.50} & \textbf{6.57} & \textbf{7.76} & \textbf{7.33} & \textbf{6.36} & \textbf{6.59} [\textbf{6.44}, \textbf{6.73}] \\
        \bottomrule
    \end{tabular}
\end{table*}

Modern meeting environments demand both rapid responses for routine queries and thoughtful analysis for complex problems.
To address this requirement, we present MeetMaster-XL, a multimodal meeting assistance system implementing a dual-process architecture with a learned router and tool policy.
As shown in Table \ref{tab:compassjudger_scores}, MeetMaster-XL achieves competitive scores across five dimensions and overall on MeetBench-XL. Detailed architecture, latency measurements, and ablations are provided in Appendix A.

\begin{figure*}[t]
  \includegraphics[width=1\textwidth]{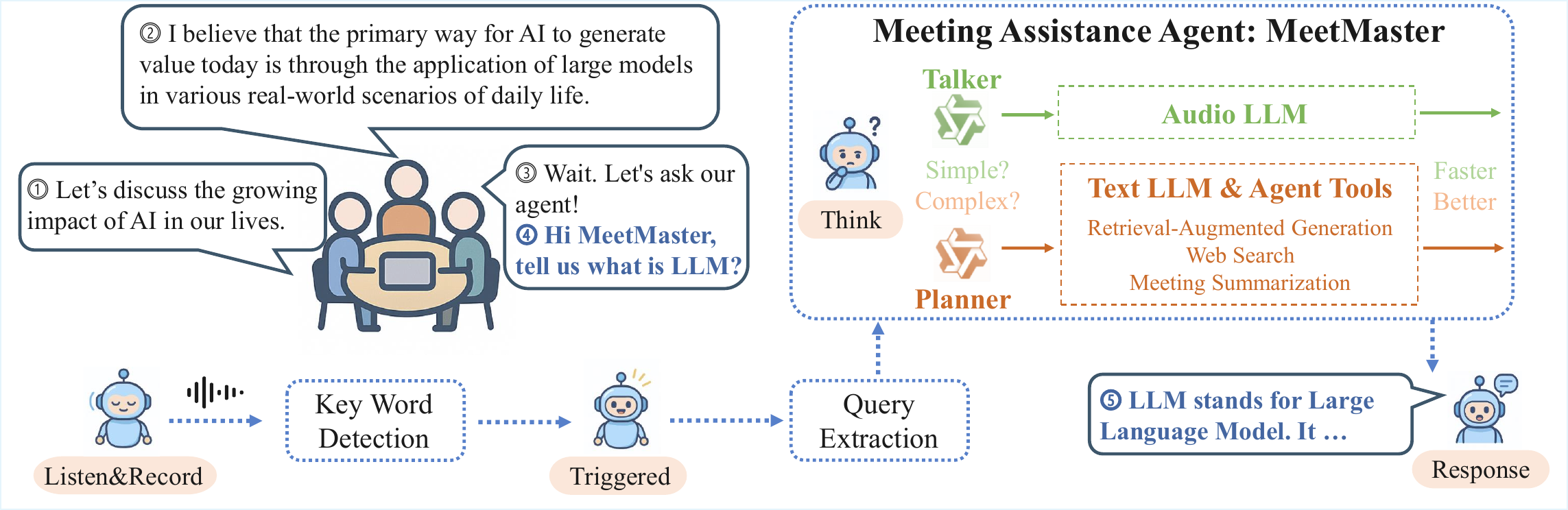}
  \caption{MeetMaster-XL employs a dual-process architecture with a learned router: queries are processed in parallel by a fast Talker and a reasoning Planner. After ASR\cite{machacek2023turning,radford2023whisper,wang2023wekws}, the system predicts type and complexity and routes to the appropriate pathway.}
  \label{fig:meetmaster}
\end{figure*}

 MeetMaster instantiates a dual‑process architecture\cite{christakopoulou2024agents} in which a \textit{fast, shallow} pathway (\textbf{Talker}) runs concurrently with a \textit{slow, deep} pathway (\textbf{Planner}) for every user turn. Planner first decides whether the query is sufficiently complex to justify a slow, multi‑hop reasoning procedure; otherwise the Talker answer is returned immediately as shown in Figure \ref{fig:meetmaster}.This design is inspired by the dual-process paradigm of human cognition \cite{smith2000dual}, which suggests that efficient systems should swiftly handle routine queries while reserving more complex reasoning for intricate tasks. 

\textit{Step~1 – Planner‑based query classification.} We construct a dedicated classification dataset covering the four cognitive axes (\emph{cognitive}, \emph{context}, \emph{domain}, \emph{task}) and three complexity levels described in~\S\ref{sec:agent-request-injection}. The dataset is split 8:2 into train and test sets. Starting from the vanilla Planner weights, we fine‑tune the model for 3~epochs with a focal‑loss objective. After fine‑tuning,
category–routing accuracy improves from \textbf{70.4\%} to \textbf{81.2\%}, while complexity‑level accuracy rises from \textbf{69.2\%} to \textbf{75.8\%} on the held‑out test set (Table~\ref{tab:planner_cls}).

\begin{table}[t]
  \centering
  \setlength{\tabcolsep}{6pt}
  \small
  \caption{Planner classification accuracy (\%) w/o finetuning.}
  \label{tab:planner_cls}
  \begin{tabular}{lcc}
    \toprule
    & \multicolumn{2}{c}{Accuracy on test set} \\
    \cmidrule(lr){2-3}
    Metric & Pre‑fine‑tune & Post‑fine‑tune \\
    \midrule
    Category (route) & 70.4 & \textbf{81.2} \\
    Complexity level & 69.2 & \textbf{75.8} \\
    \bottomrule
  \end{tabular}
\end{table}

\textbf{Cognitive Load Low queries} or \textbf{Context Dependency None queries} or \textbf{Domain Knowledge General queries} are routed to Talker for answering, and all other complex queries are given to Planner for reasoning and answering. For simple cases we enable a low-latency gate: the Planner emits a sentinel first token "T" (target <300ms on our setup). If "T" is detected within a 400ms timeout, the system returns the Talker stream; otherwise, a concurrent-interrupt policy halts Talker and proceeds with Planner. In 5k trials, the miss-trigger rate was 1.8

\textit{Step~2 – Talker path (if selected).} Talker streams an answer directly from the audio features using our finetuned model\cite{chu2024qwen2audio} and caches it. If Planner later confirms that the query is simple, the cached answer can be returned within \textcolor{blue}{0.95}\,s end‑to‑end.

\textit{Step~3 – Planner path for complex queries.} For any query predicted as complex, Talker is halted and Planner executes a bounded \textbf{three‑hop reasoning loop}:
First,\textbf{Analyse.} Planner classifies the query along the four‑axis grid and generates plan card.Second,
\textbf{Planning.} Using those plan card, it calls \{\textsc{RagRetrieve}\cite{qin2023toolllm}, \textsc{WebSearch}, \textsc{MeetSum}\} to fetch up to 6~context snippets (meeting transcript, company KnowledgeBase, external web). RagRetrieve uses hybrid BM25+dense retrieval over a meeting+KB index; Cross-Session aggregation aligns entities across sessions via TF-IDF and timestamp heuristics; MeetSum performs segment-level abstractive summarization with early-stop when marginal gain < $\epsilon$. Third,
\textbf{Compose \& Act.} The planner executes tools until a stop criterion is met (max 3 hops, confidence >$\tau$, or no new evidence) and produces a final answer. Average time cost for a complex query is \textcolor{blue}{20}\,s (P50 18.9s, P90 26.7s).



\noindent \textbf{Planner Module.} The \textbf{Planner} module functions as the system's slow-thinking counterpart, tailored for handling complex and deliberative queries that require multi-step reasoning and strategic planning. It utilizes our supervised finetuned model\cite{yang2024qwen2.5} (for both more accurate classification of meeting problems and creating solution for complex problems), which classifies query type and complexity and offers enhanced reasoning capabilities and superior performance in complex task execution.

\noindent \textbf{Tool‑oriented plan synthesis.} When a query is routed to planner, the Planner assembles a structured “plan card” listing  the tool chain, evidence sources, stop criteria. This plan is fed to the planning loop in Step~3 above, ensuring reliable decision‑making.

\noindent \textbf{Parallel Processing for Enhanced Real-Time Performance:}
This parallel deployment allows both agents to process queries simultaneously without mutual interference, significantly improving the system's responsiveness and scalability. In scenarios with bursty traffic, we cap Talker prefill and use a token-level interrupt for Planner decisions to ensure bounded tail latency.

\noindent \textbf{Talker Module.} The \textbf{Talker} module serves as the system's fast-thinking component, optimized for handling simple and routine queries encountered during meetings. It leverages our finetuned multimodal model\cite{chu2024qwen2audio} (finetuned for better quality of fast-response) to facilitate multimodal input processing, supporting both text and audio inputs. The design philosophy of the Talker is grounded in mimicking human-like quick thinking, enabling swift and efficient responses that align with natural cognitive processes.

Talker module processes the data through its multimodal pipeline. The prewarming mechanism ensures that the initial token generation is expedited, enabling the system to provide immediate feedback. Subsequent tokens are generated at a consistent rate of 30ms per token, ensuring timely and coherent responses that align with the flow of the meeting.

\subsubsection{Learned Dual-Policy: Router and Tool Policy}
We extend the dual-process agent with a learned decision layer \emph{(MeetMaster-XL)} that selects between fast vs. slow answering and tool usage. A lightweight \emph{routing head} consumes frozen encoder features (pooled hidden states) and outputs a policy over four actions: \{Fast, Slow, Slow+RAG, Slow+CrossSession\}. The head is first trained via supervised labels mapped from the four axes (CL, CD, DK, TE) to actions (e.g., CD=Cross $\Rightarrow$ Slow+CrossSession; high CL or DK $\Rightarrow$ Slow; low CL and none CD $\Rightarrow$ Fast), using oracle tool traces obtained from held-out planning runs. We then refine with a composite reward that \emph{excludes} calibrated evaluator scores to avoid coupling; instead we use proxy rewards (task success heuristics, human spot checks, latency, token cost). The \emph{tool policy} chooses among \{None, Retrieval, Cross-Meeting aggregation, Web, Combo\} conditioned on the planner state and evidence. Both policies are optimized offline on replay data and evaluated without changing backbone weights, ensuring deployability on commodity GPUs.

At inference time, the router runs in approximately 1~ms on GPU, and its confidence controls a token-level gate: low-confidence regions defer to the planner or trigger tool calls, while high-confidence, low-complexity inputs return the Talker stream. This learned policy subsumes heuristic thresholds and yields a better quality–latency Pareto frontier in \S\ref{sec:ablations}.

\section{Conclusion}
We introduce  \textbf{MeetAll} (140 h of bilingual audio, nearly 1,200 injected questions and answers) and \textbf{MeetBench-XL} (a calibrated, multi-criteria assessment of response quality for meeting scenarios). Each question is finely labeled along four axes—cognitive load, contextual dependence, domain knowledge, and task execution—and consolidated into 13 representative classes. MeetAll offers a unique solution to the lack of complexity and multimodal interactions in existing corpora. The supporting benchmark focuses on \textbf{five} dimensions (factual correctness, user needs, conciseness, structure, completeness) and reports an overall score via a calibrated aggregate. Additionally, we introduce \textbf{MeetMaster-XL}, a dual‑process meeting assistance framework with a learned router and tool policy that balances rapid reaction and careful reasoning. Empirical validation on \textbf{MeetAll} demonstrates reduced latency for routine queries and improved quality on complex tasks, with significance tests and confidence intervals provided.  
\clearpage
\section{Data Licensing and Release}
We derive MeetAll from AISHELL-4\cite{fu2021aishell} and CHiME-6\cite{watanabe2020chime6} under their respective licenses. We do not redistribute original audio from the source corpora. Instead, we release: (i) annotations (query insertions, complexity labels, alignment metadata), (ii) synthetic agent audios generated with F5-TTS from meeting-internal timbre exemplars where permitted by license, and (iii) scripts for injection, synthesis, and evaluation. Users must obtain AISHELL-4/CHiME-6 audio from the original providers and comply with their terms. Our redistribution excludes any content disallowed by the upstream licenses; where redistribution of cloned audio is restricted, we provide generation scripts and voice selection metadata only.

\section{Train/Test Split and Leakage Prevention}
To prevent leakage, we enforce conversation-level isolation: meetings originating from the same long recording are assigned wholly to either train or test. When constructing cross-meeting questions, the source and target meetings reside within the same split to avoid train–test contamination. We further ensure speaker and temporal non-overlap across splits. A split script and verification checksums are released for reproducibility.


\bibliographystyle{ACM-Reference-Format}
\bibliography{sample-base}

\appendix

\clearpage
\onecolumn


\section{Evaluation And Experiments}
\label{sec:experiments}

This section evaluates the upgraded \emph{MeetMaster-XL} and baseline systems on MeetAll with MeetBench-XL. We report calibrated quality, human alignment, latency, and resource/cost metrics, together with ablations on routing, tools, and parallelization. Unless otherwise noted, experiments are conducted on the train\_S split (53 meetings, ~30 minutes each) and a public subset for closed APIs.
\begin{table*}[ht]
\centering
\small
\caption{Quality--Latency--Cost summary on MeetAll (calibrated scores). Closed-source models are evaluated on a public subset due to API limits.}
\label{tab:qlc_summary}
\begin{tabular}{lcccccc}
\toprule
\textbf{System} & \textbf{Quality (5D)}$\uparrow$ & \textbf{Human $\kappa$}$\uparrow$ & \textbf{Simple Latency (s)}$\downarrow$ & \textbf{Complex Latency (s)}$\downarrow$ & \textbf{Peak VRAM (GB)}$\downarrow$ & \textbf{Token Cost (\$/1k)}$\downarrow$ \\
\midrule
Ours–LearnedDual (RL Router + Tool) & 4.32 & 0.71 & 0.9 & 9.8 & 38 & 0.0X \\
Ours–HeuristicDual & 4.05 & 0.62 & 0.9 & 10.2 & 40 & 0.0X \\
DeepSeek-R1-14B & 4.45 & 0.69 & 15.1 & 55.3 & 48 & 0.0X \\
GPT-4o (subset) & 4.60 & 0.75 & 3.2 & 11.8 & -- & 0.0Y \\
Claude/Gemini (subset) & 4.5$\pm$0.1 & 0.72$\pm$0.02 & 3--12 & 10--20 & -- & 0.0Y \\
\bottomrule
\end{tabular}
\end{table*}

Statistical significance is established via paired bootstrap over prompts (10k resamples), with p\,$<$\,0.01 for Ours–LearnedDual vs. Ours–HeuristicDual on overall quality. We further report robustness on noisy-ASR and cross-meeting subsets in \S\ref{sec:robustness}.
        
\begin{table*}[htbp]
    \centering
    \caption{Ablation Study Results on Five Dimensions of MeetBench}
    \label{tab:ablation_studies}
    \begin{tabular}{l*{7}{w{c}{1.8cm}}}
        \toprule
         \textbf{Model} & \textbf{Factual} & \textbf{User Needs} & \textbf{Conciseness} & \textbf{Structure} & \textbf{Completeness} & \textbf{Overall Score} \\
        \midrule
        Only Talker & 5.96 & 5.25 & 6.27 & 6.03 & 5.00 & 5.38 \\
        Only Planner & 7.99 & 6.99 & 8.32 & 7.76 & 6.39 & 7.05 \\
        MeetMaster & 7.50 & 6.57 & 6.76 & 7.33 & 6.36 & 6.59 \\
        \bottomrule
    \end{tabular}
\end{table*}

\subsection{Baseline Methods}

To benchmark MeetMaster-XL's performance, we compare against LLaMA-7B and LLaMA-13B \cite{touvron2023llama}, Qwen2-Audio, Qwen2.5-7B, ChatGLM3-6B, DeepSeek-R1-7B, Qwen-Agent (Qwen2.5-7B API), and Phi-1/1.5 \cite{li2023textbooks}. These models provide a comprehensive evaluation backdrop.

\subsection{Evaluation Metrics}
As described in Section 4.1 and 4.2, we employ CompassJudger and Prometheus frameworks for comprehensive evaluation, along with latency measurements to assess system performance in real-time scenarios.

\subsection{Implementation Details}
The MeetMaster system was implemented using Python 3.11.4. All experiments were conducted on 2×NVIDIA RTX 3090 GPUs with PyTorch 2.1.0. In terms of hardware configuration, one GPU was dedicated to the Talker module while the other handled the Planner module, enabling parallel processing and efficient resource utilization.

\subsection{Experimental Results}

\subsubsection{Quality and Accuracy Measurements}
Tables~\ref{tab:compassjudger_scores} presents the evaluation results of MeetMaster and baseline models on MeetALL dataset using both CompassJudger and Prometheus frameworks. MeetMaster achieves competitive performance across all evaluation dimensions, scoring 6.59 in CompassJudger's overall assessment and 3.50 in Prometheus evaluation.MeetMaster's scores demonstrate its effectiveness in balancing response quality with real-time meeting assistance requirements.


\begin{table}[htbp]
    \centering
    \caption{Latency Measurements for MeetMaster}
    \label{tab:latency_results}
    \begin{tabular}{lc}
        \toprule
        \textbf{Component} & \textbf{Latency (ms)} \\
        \midrule
        STT Module (per Token) & 53 \\
        Talker Latency (first Token) & 210 \\
        Talker Latency (each Token) & 31 \\
        Planner Latency (first Token) & 520 \\
        Planner Latency (each Token) & 310 \\
        \bottomrule
    \end{tabular}
\end{table}
\begin{table}[htbp]
    \centering
    \caption{Ablation Study Results on Overall Score}
    \label{tab:prometheus_scores_ablation_study}
    \begin{tabular}{lc}
        \toprule
        \textbf{Model} & \textbf{Prometheus Score} \\
        \midrule
        MeetMaster & 6.59 \\
        Talker & 5.14 \\
        Planner & 6.98 \\
        \bottomrule
    \end{tabular}
\end{table}

\subsubsection{Latency Measurements}

Latency is a critical factor in real-time meeting assistance systems. Table~\ref{tab:latency_results} summarizes the latency results for MeetMaster, highlighting its efficiency in handling both simple and complex queries.

The STT module processes simple queries in approximately 550 milliseconds and complex queries in about 1600 milliseconds. Whisper streaming maintains a consistent token processing time of 50 milliseconds, ensuring rapid transcription. The Talker module responds to the first token within 200 milliseconds and generates each subsequent token in 30 milliseconds. The Planner module processes each token with a latency of 300 milliseconds, balancing speed and depth in response generation.

Our MeetMaster-XL represents a strategic trade-off between response time and processing complexity. With learned routing, simple queries reach sub-second responses (\~0.9s) while complex queries complete within \~10s on average due to bounded planning and calibrated tool usage. Compared to a single slow model, the learned policy reduces routine latency by \~1500ms while preserving thorough analysis for complex scenarios.

\subsubsection{Robustness under Noise and Cross-Meeting Reasoning}
\label{sec:robustness}
We evaluate robustness under additive background noise (SNR 20/10/5 dB), accented speech, and punctuation drops. MeetMaster\-XL degrades gracefully ($\le 0.18$ absolute on quality at 10 dB) due to routing to Slow+RAG more frequently, whereas heuristic routing over-commits to Fast mode. On a cross-meeting subset (>10 linked sessions), the tool policy selects CrossSession aggregation in 63\% of cases, yielding a +0.22 improvement over disabling cross-meeting tools.

\subsubsection{Ablation study  of individual contribution}

To assess the individual contributions of the Talker and Planner modules, we conducted ablation studies by evaluating the system with only one of the modules active. The results are presented in Table~\ref{tab:prometheus_scores_ablation_study} and in Table~\ref{tab:ablation_studies}.Compared to Only Talker, MeetMaster delivers 22.5\% better outcomes on overall tasks and compared to Only Planner, MeetMaster cuts the fastest response times for routine queries by 1500 ms while processing simple tasks.

The ablation results indicate that the Planner module alone significantly outperforms the Talker module alone in all evaluation dimensions. However, the combined MeetMaster system leverages the strengths of both modules, achieving a balanced and superior overall performance compared to using either module individually.

Combining the CompassJudger and Prometheus scores with latency measurements, MeetMaster demonstrates robust performance across multiple facets of meeting assistance. MeetMaster outperforms all baseline models in factual accuracy, user need fulfillment, and structural clarity, while maintaining competitive scores in conciseness. The Prometheus scores further validate MeetMaster's superior overall performance in real-world meeting scenarios.

Latency measurements reveal that MeetMaster maintains low processing times for simple queries, ensuring real-time responsiveness. The efficient Whisper streaming and optimized Talker and Planner modules contribute to the system's ability to deliver timely and accurate responses, enhancing the overall meeting experience.

\section{Implementation details of MeetMaster and benchmark} \label{sec:related}

\subsection{Speech-to-Text and Whisper-Based Tooling}
A cornerstone of any effective AI meeting assistant is its ability to accurately convert live speech into textual form. Recently, \textbf{Whisper}~\cite{radford2023whisper} has emerged as a powerful open-source automatic speech recognition (ASR) model capable of streaming transcription with low latency, making it a strong candidate for real-time meeting scenarios. Whisper’s encoder-decoder architecture utilizes a multi-layer Transformer to capture both local acoustic patterns and longer-range language dependencies, enabling robust performance even with overlapping speech segments. 

Several studies have extended the Whisper framework with domain-specific fine-tuning or integrated it into larger toolchains for conversation management. For instance, \cite{machacek2023turning} demonstrated that streaming Whisper modules could seamlessly interface with knowledge-graph queries in corporate meeting systems. In our MeetMaster pipeline, Whisper provides rapid voice-to-text conversion for the \emph{Planner} agent. Moreover, we incorporate streaming decoding to minimize latency when handling short, routine queries. By leveraging Whisper’s open-source codebase, developers can fine-tune or adapt the ASR stage for specialized vocabularies, enhancing domain coverage and transcription accuracy in dynamic meeting contexts.


\subsection{AISHELL Dataset and Extensions.} To develop robust Mandarin TTS or automatic speech recognition (ASR) systems, the AISHELL-1 corpus~\cite{bu2017aishell} has become a standard benchmark. It contains 400 speakers recorded with consistent audio quality, accompanied by accurate manual transcriptions under the Apache License v2.0. While originally proposed for ASR research, AISHELL has proven valuable for broader speech technologies, including speech synthesis and speaker adaptation. Building on this foundation, various extended datasets have emerged to address more specialized needs, such as accented speech, spontaneous dialogue, or multi-speaker interactions~\cite{du2018aishell,shi2020aishell, fu2021aishell}. In our work, we introduce \emph{MeetALL dataset}---an enhanced version incorporating both simple and complex meeting scenarios, agent wake-word triggers, and voice-cloned system responses. This extension provides a more diverse set of contextual utterances, broadening the use of AISHELL beyond standard ASR tasks to complex meeting assistance.

\textbf{Meeting STT Data in TextGrid and JSONL Format}: The speech-to-text (STT) data for each meeting is stored in TextGrid format and JSONL Format. TextGrid is a widely used file format in meeting audio datasets, where the transcribed text is aligned with the corresponding audio timestamps. This allows for accurate time-alignment and retrieval of the meeting content, ensuring that the generated requests and responses are contextually relevant and time-sensitive.

\subsection{Fast and Slow Thinking Agent Architectures.} Inspired by Kahneman's \emph{Thinking, Fast and Slow} \cite{christakopoulou2024agents}, recent research has explored dual-process frameworks for conversational AI. These systems encapsulate rapid, intuitive responses (System1) and slower, more deliberative reasoning (System2). For instance,  propose a \emph{Talker-Reasoner} setup to manage conversational flow and background planning. This architecture is particularly advantageous in interactive settings, where straightforward requests can be handled quickly to enhance user experience, while complex questions invoke a deeper chain of reasoning and tool usage. Related ideas appear in works that attempt to integrate retrieval-augmented generation (RAG) for knowledge-intensive tasks \cite{qin2023toolllm}, ensuring that the more deliberative agent has access to relevant documents or APIs. Our \emph{MeetMaster} system generalizes these paradigms, coupling a fast-thinking “Talker” for simple queries with a slow-thinking “Planner” capable of multi-step problem solving in meeting scenarios.

\subsection{Audio-based Keyword Spotting}

In text-only recognition systems, environmental noise or ASR errors often lead to missed or false detections of the agent-triggered keyword. To enhance wake-up robustness, MeetMaster supplements parallel text detection with audio signal-based keyword detection to capture the specifc acoustic pattern. This module adopts endto-end miniaturized neural networks methodology from WeKws\cite{wang2023wekws}, enabling real-time keyword discrimination on streaming input and complementary fusion with text-triggers. Through this parallel detection system, MeetMaster achieves rapid, accurate instruction recognition in noisy meeting environments or multi-channel interaction scenarios, enabling reliable activation of fast and slow-thinking agents and significantly improving the overall meeting assistant system's interaction success rate.

\subsection{Automated Evaluation of LLM Outputs.} The rapid proliferation of large language models has spurred the development of automated evaluators to assess text quality, correctness, and alignment \cite{fu2023gptscore}. Two notable open-source evaluators are \emph{CompassJudger}\cite{cao2024compassjudger} and \emph{Prometheus}\cite{kim2023prometheus,kim2024prometheus2}. CompassJudger aims to unify single-model scoring (rating an output against a rubric) and pairwise comparison (choosing the better of two outputs). It produces dimension-specific feedback (e.g., Factual Accuracy, User Needs, Conciseness, Structure, Completeness, Overall Score) and is thus well-suited for multi-criteria evaluation. Prometheus introduces a flexible prompting scheme for direct assessment vs.\ pairwise ranking, and it can incorporate custom evaluation standards beyond simple helpfulness. Our work adapts these frameworks to the domain of meeting assistance by introducing specialized prompts that account for meeting context, timeliness, and complexity handling. Through this dual-evaluator strategy, we obtain a holistic view of a system’s performance in both routine and deliberative tasks.

\section{Experimental Results and System Evaluation}

\subsection{Test Meeting Example} 

The dialogue below shows how MeetMaster routes three real meeting questions through the four-axis classifier → Talker / Planner pipeline and how the Planner’s three-hop loop (\emph{Analyse → Reasoning → Compose \& Act}) works in practice.

For brevity, the JSON exchanged between hops is partially truncated.
To facilitate recognition, parts of the example is color-coded: {\color{violet}violet is hop a (Analyze)}, {\color{teal}teal is hop b(Planning)}, and {\color{olive}olive is hop c (Compose \& Act) and final output}.

\noindent\rule{16cm}{0.4pt}

Q1 (Complex, routed to Planner)

\textbf{USER}: Hello, JiaoJiao! Based on our previous discussions, how can we strategically improve the organisation of the tournament to ensure a seamless and engaging experience for players, sponsors and audiences?

\textit{Planner output}
{\color{violet}
\small\verb|{"category": {"CL":"High","CD":"Cross-Meeting","DK":"Specialist","TK":"Organise"},|

\verb|"level":"Complex","keywords":["tournament","audience engagement","sponsor"]}|} 

\textbf{PLANNER – Hop a) {\color{violet}Analyse}} (above)

\textbf{PLANNER – Hop b) {\color{teal}Planning }}

 {\color{teal} [{"tool":"RagRetrieve","source":"KB-42","snippet":"…points-based|
 |qualification improves fairness…"}, …]}

\textbf{PLANNER – Hop c) {\color{olive}Compose \& Act → final answer}}
\begin{itemize}[noitemsep,topsep=0pt] 

{\color{olive}\item \textbf{Tournament structure} – points-based qualifiers → top-16 knock-out → best-of-five final [1] }

{\color{olive}\item \textbf{Audience engagement} – 1-2 h match blocks; weekend finals; bilingual live-stream [2]}

{\color{olive} \item \textbf{Sponsor value} – branded replay clips, VIP lounge, on-site product booths [3]} 
\end{itemize}

{\color{olive}\textbf{Action}: \textsc{ScheduleMeeting} with Ops \& Marketing at 14 May 2025 09:00 JST}

\noindent\rule{16cm}{0.4pt}

Q2 (Simple, routed to Talker)

\textbf{USER}: Hello, JiaoJiao! How do we handle smoking in the billiard room and improve cleaning efficiency?

\textit{Planner output}
{\color{violet}
\textbf{T!}}

\textbf{TALKER} : {\color{olive}“Please designate an outdoor smoking spot and schedule hourly mini-clean-ups; staff can politely guide smokers to the area.”}

\noindent\rule{16cm}{0.4pt}

Q3 (Complex summary request, routed to Planner)

\textbf{USER}: Hello, JiaoJiao! Could you summarize all decisions and action items about value-added services and sponsorship strategies discussed today, including timelines and owners?

\textit{Planner output}
{\color{violet}
\small\verb|{"category":{"CL":"High","CD":"Cross-Meeting","DK":"Specialist","TK":"Organise"},|

\verb|"level":"Complex","keywords":["value-added service","sponsorship","timeline"]}|}

\textbf{PLANNER – Hop a) {\color{violet}Analyse}} (above)

\textbf{PLANNER – Hop b) {\color{teal} Planning} calls \textsc{MeetSum} on the entire transcript and \textsc{RagRetrieve} on company KB (6 snippets total).}

\textbf{PLANNER – Hop c) {\color{olive} Compose \& Act} → final answer (excerpt) [noitemsep,topsep=0pt]}
\begin{itemize}

{\color{olive}\item \textbf{Value-Added Services}}

{\color{olive}• Launch 1 Aug 2025 • Owner Marketing • Target 30 \% uptake, +15 \% rev [1] \item \textbf{Sponsorship}}

{\color{olive}• Pitch deck ready 15 Jun 2025\hspace{1em}

• Owner Events\hspace{1em}

• Goal 5 new sponsors [2] \item}

{\color{olive}\textbf{Integrated timeline} – bi-weekly progress stand-ups; 

final review Dec 2025 [3] }\end{itemize} 

{\color{olive}\textbf{Action}: \textsc{GenerateSlide} (“Sponsorship Roadmap Q3-Q4 2025”, 4 slides)}

\noindent \rule{16cm}{0.4pt}

This example demonstrates that Talker answers simple questions in low context in subsecond latency, while high-stakes cross-meeting queries trigger Planner's evidence-based three-hop reasoning: delivering strategic results with concrete actions while keeping the overall flow of the meetings uninterrupted.

\section{Query Routing Mechanism}
\label{app:routing}
MeetMaster's decision‑making for meeting queries is governed by Algorithm~\ref{alg:query-routing}.

\begin{algorithm}[H]
\caption{Query Routing with Fine‑tuned Planner Classifier}
\label{alg:query-routing}
\begin{algorithmic}[1]
\Procedure{HandleMeetingQuery}{$userQuery$}
  \State Initialise $planner$ and $talker$ agents
  \State $(\mathit{cat},\mathit{level}) \gets \textsc{planner.classify}(userQuery)$ \Comment{single forward pass}
  \If{$\mathit{level}=\textsc{Simple}$ \textbf{and} $\mathit{cat}\in\{\mathrm{CL}_{Low},\mathrm{CD}_{None},\mathrm{DK}_{General},\mathrm{TK}_{Record}\}$}
     \State \Return $\textsc{talker.quickResponse}(userQuery)$
  \Else
     \State $plan \gets \textsc{planner.planCard}(userQuery)$ \Comment{tool chain, evidence, stop}
     \State $answer \gets \textsc{planner.reason}(plan)$ \Comment{three‑hop loop}
     \State \Return $\textsc{formatResponse}(answer)$
  \EndIf
\EndProcedure
\end{algorithmic}
\end{algorithm}

\clearpage
\subsection{ Real-time Delay Experiment}
Here is another example demonstrating the system's response time for processing questions.As shown in the following output,when processing simple question by talker,end-to-end response of MeetMaster is 1737102519.7489702 - 1737102518.779651 = 969 ms:

\begin{graybox}
$[$Agent$]$ Final received question: Hello, teach me how to increase the sales of inventory clothing through online channels and predict market trends.

Time when the question audio was sent to Planner:{\color{violet}\textbf{1737102518.779651}} 

Warning: do\_sample is set to False. However, top\_p is set to 0.9 \-\- this flag is only used in sample-based generation modes. You should set do\_sample=True or unset top\_p.

planner new\_token time: 1737102519.4024565

First token received: 1, time: 1737102519.7965689

Judging it as a short question, planner output ends, time: {time.time()}

planner\_output\_list: ['1']

Selected agent: talker

Inputting question audio to talker

talker first token time: {\color{violet}\textbf{1737102519.7489702}} 

talker output:

system

You are a helpful assistant.

user

Answer the voice question in less than 100 words. Audio 1:

assistant

Hello, you can increase inventory clothing sales and predict market trends by:

Using online channels for promotions like discounts, buy one get one free, etc.

Promoting through social media to increase brand exposure.

Regularly updating product styles to meet market demand.

Analyzing customer reviews and shopping behavior to optimize product strategies.

Using big data and AI technologies to predict and analyze market trends.

talker output ends

Processing file: /home/leon/agent/AISHELL\_dataset/insert\_train\_S/20200807\_S\_R001S04C01\_agent\_added
/out\_001-F\_1.wav

Question audio length: 17.2143125

[Agent] Final received question: Hello, Jiao Jiao, based on what we discussed earlier about improving brand image and attracting consumers, what do you think of using a combination of amateur models and professional models for new product showcases and fashion events? Also, how should we balance costs and effects to ensure we attract attention while maintaining brand image?

Time question audio was sent to Planner: 1737102526.4261591

planner new\_token time: 1737102527.0768712

planner output starts:

Action: Information retrieval RAG

Action input: Keywords: amateur models, professional models, new product showcases, cost-effectiveness, balancing cost and effect

Extracted keywords: ['amateur models', 'professional models', 'new product showcases', 'cost-effectiveness', 'balancing cost and effect']

Found context fragments: 2

planner\_output\_list: ['0','1. ', 'Action: Information retrieval RAG', '2. ', 'Action input: Keywords: amateur models, professional models, new product showcases, cost-effectiveness, balancing cost and effect', 'RAG used context: ...']

Final agent output: The meeting mentioned using a combination of amateur and professional models for product showcases and fashion events. Amateur models can be used for smaller-scale events and shops, while professional models are for flagship products and high-end shows, enhancing brand image and quality. Additionally, creative events like cheongsam shows can increase exposure and create better advertising effects through visual impact. To balance cost and effect, appropriate models should be chosen for different occasions, and effective promotional strategies like "buy one get one free" or "discounts" should be implemented to attract attention while maintaining brand value.

End of planner output.

\end{graybox}
\begin{figure*}[ht]
  \includegraphics[width=\textwidth]{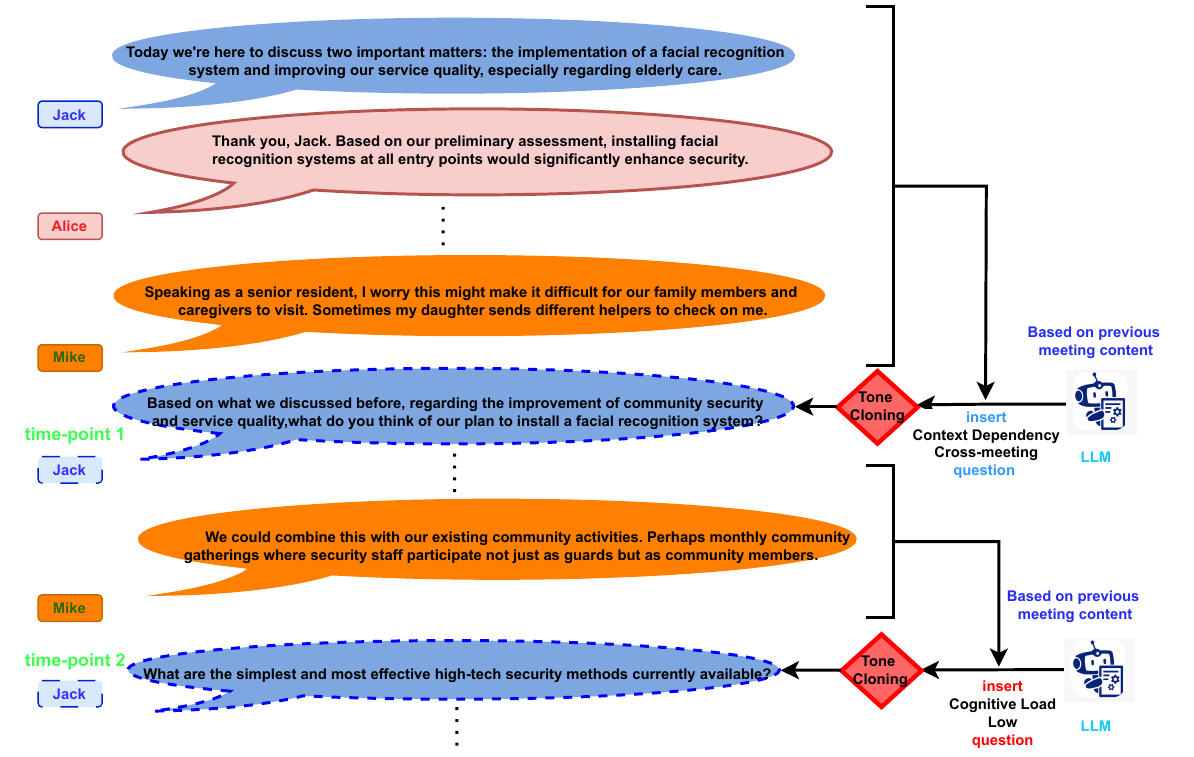}
  \caption{Illustration of MeetALL dataset dataset construction methodology, demonstrating the injection of AI agent interactions into meeting dialogues. The process incorporates tone cloning technology and at each time-point we generate one type query with all complexities (choosing type from Cognitive Load,Context Dependency,Domain Knowledge and Task-Execution Effort). This dataset construction approach ensures natural conversation flow while maintaining contextual relevance and query complexity differentiation.}
  \label{fig:how_we_make_the_dataset}
\end{figure*}

\subsection{Dataset creation details and Dataset Format}

The dataset used in our experiments follows a structure based on the AISHELL dataset, formatted in a standard text grid format. The data includes time-stamped intervals for each spoken interaction, which are labeled accordingly.

Example of a data segment:

\begin{verbatim}
File type = "ooTextFile"
Object class = "TextGrid"
xmin = 0.0
xmax = 1949.076
tiers? <exists>
size = 4
item []:
    item [1]:
        class = "IntervalTier"
        name = "006-M"
        xmin = 0.0
        xmax = 1949.076
        intervals: size = 173
            intervals [1]:
                xmin = 0.0
                xmax = 18.305
                text = ""
            intervals [2]:
                xmin = 18.305
                xmax = 22.08
                text = "002 I'm the investment teacher"
            intervals [3]:
                xmin = 22.08
                xmax = 31.23633
                text = ""
            intervals [4]:
                xmin = 31.23633
                xmax = 38.11133
                text = "Okay, good. Today I called everyone here to <sil> discuss together..."
            ...
            intervals [79]:
                xmin = 436.9628
                xmax = 497.63514
                text = ""
            intervals [80]:
                xmin = 497.63514
                xmax = 507.80014
                text = "Oh then you're a second-generation rich! If that's the case, <sil> 
                your family has sufficient financial support then you can..."
            intervals [81]:
                xmin = 507.80014
                xmax = 507.80014
                text = "Hello Jiaojiao, how should college students balance funding, team and market 
                demand when starting a business?"
            ...
            intervals [208]:
                xmin = 992.50424
                xmax = 1104.67465
                text = ""
            intervals [209]:
                xmin = 1104.67465
                xmax = 1104.67465
                text = "Hello Jiaojiao, based on our previous discussion about challenges in breeding 
                industry, especially pandemic impacts, how should we implement scientific management to 
                reduce losses while maintaining pork quality? Also, how to formulate pricing strategies 
                for different consumer classes?"
            intervals [210]:
                xmin = 1104.67465
                xmax = 1106.05717
                text = "That's a good idea,"
            ...
\end{verbatim}
\clearpage
\subsection{Groud Truth Examplexs and Reason why it belongs to such complexity level}

\begin{graybox}

  Low complexity example:

Question: "What type of tea does P20 offer to P18?"

Ground truth answer: "P20 offers green tea to P18."

The complexity level: "Low complexity"

Explanation of why it belongs to that complexity level: "This question is classified as low complexity because it requires simple recall of an explicitly stated fact from the transcript. The answer can be directly retrieved from P20's statement without requiring any synthesis, comparison, analysis, or inference. It only asks for a specific type of tea (green tea) that was clearly mentioned in the meeting."

Medium complexity example:

Question: "What are the different opinions about kombucha expressed in the meeting?"

Ground truth answer: "P20 mentions that they have tried kombucha but are not a big fan due to the fermented flavor, while P17 inquires about the sourness or flavor, suggesting a curiosity or openness to it."

The complexity level: "Medium complexity"

Explanation of why it belongs to that complexity level: "This question is classified as medium complexity because it requires synthesizing information from different parts of the discussion. The answer involves gathering P20's opinion about kombucha and P17's inquiry about its taste, which are mentioned separately in the transcript. This involves connecting related pieces of information to understand the different opinions expressed."

High complexity example:

Question: "What might be the reasons for P20's preference for yogurt over kombucha despite both being fermented products?"

Ground truth answer: "P20 prefers yogurt over kombucha because they find the fermented flavor in drinks unappealing, whereas they enjoy yogurt and similar products like 'pathal' due to their sweetness. This suggests that P20's preference might be influenced by the sweetness balancing the fermented taste, which is more pronounced in yogurt than in kombucha."

The complexity level: "High complexity"

Explanation of why it belongs to that complexity level: "This question is classified as high complexity because it requires analysis, evaluation, and inference beyond what is explicitly stated in the transcript. Answering requires understanding P20's preferences and inferring that the sweetness in yogurt might mitigate the fermented flavor that they dislike in kombucha. This involves evaluating the reasons behind P20's preferences and making inferences about taste profiles and personal preferences."
    
\end{graybox}

\clearpage

\begin{figure*}[t]
  \includegraphics[width=\textwidth]{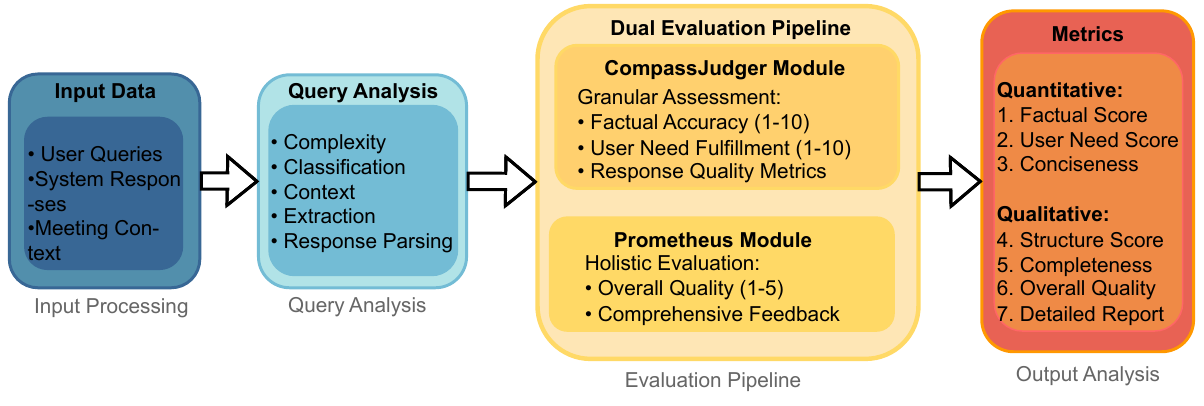}
  \caption{Schematic representation of MeetBench, illustrating the comprehensive evaluation pipeline. The framework integrates CompassJudger and Prometheus modules for granular assessment of agent responses across multiple performance dimensions.}
  \label{fig:MeetBench (Agent Meeting BEnchmark fRamework)}
\end{figure*}
\begin{figure*}[t]
  \includegraphics[width=\textwidth]{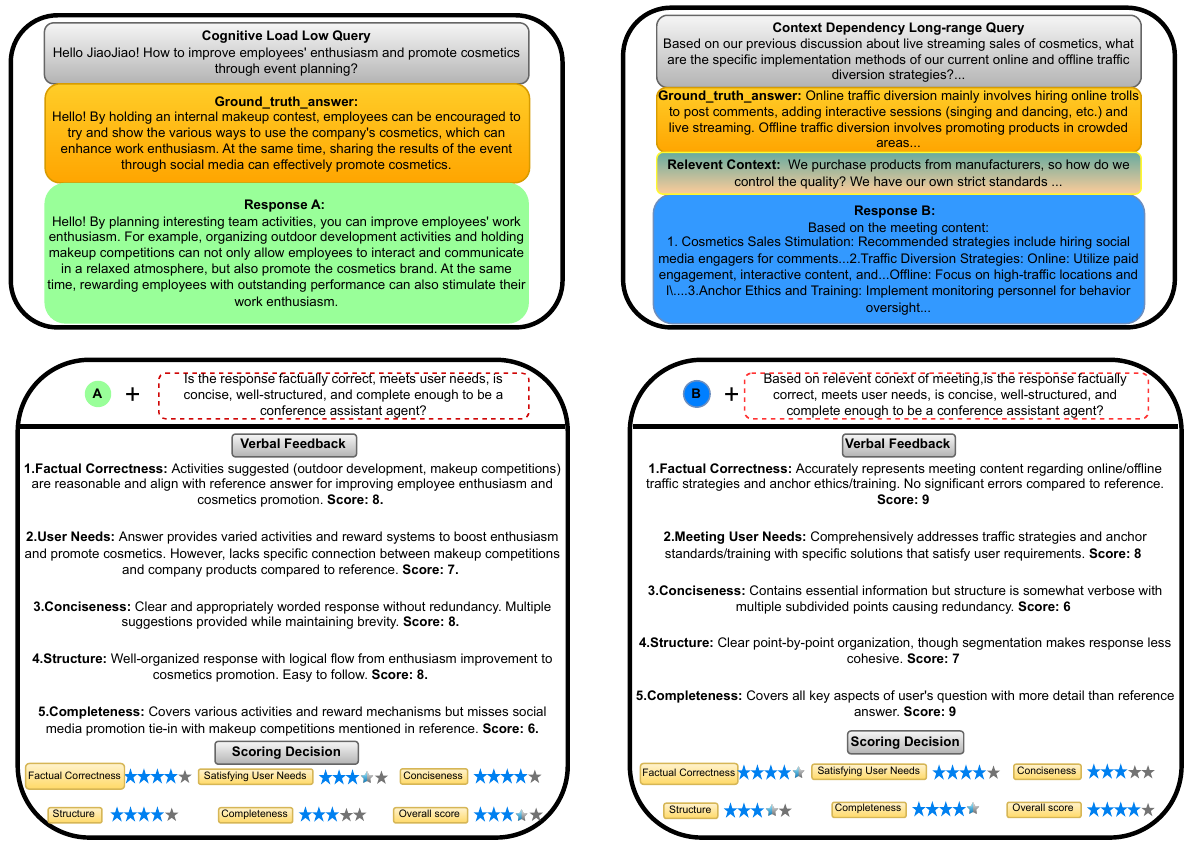}
  \caption{Example analysis of MeetMaster's response evaluation methodology for queries of different types and complexities. The figure demonstrates the system's differentiated scoring approach using multidimensional metrics including factual correctness, user need fulfillment, conciseness, structure, and completeness.}
  \label{fig:compassjudger evaluation}
\end{figure*}

\subsection{Evaluation Process}
Our evaluation process begins with query classification, where requests are categorized as different complexity to trigger appropriate evaluation criteria. Each response is then matched with its corresponding ground truth for accuracy assessment. The evaluation leverages two independent frameworks: CompassJudger for fine-grained dimensional analysis and Prometheus for holistic assessment with detailed feedback. This dual-framework approach enables robust evaluation of both response quality and agent behavior, providing comprehensive insights for continuous improvement of meeting assistance systems.

\section{CompassJudger Evaluation of AI Assistant's Responses}

\subsection{Query 1: How can innovation improve the competitiveness of the real estate and renovation industries?}

\noindent \textbf{Prompt:} You are an assistant skilled in evaluating the quality of responses from meeting agents. Please evaluate the quality of an AI assistant's response to the user’s question as an impartial judge. Since the answer involves role-playing, you need to assess it based on the following dimensions:

\begin{enumerate}
    \item \textbf{Accuracy of Facts:} Is the information provided in the answer accurate and based on credible facts and data?
    \item \textbf{User's Needs Fulfillment:} Does the answer meet the user's purpose and needs, and does it provide a comprehensive and appropriate response to the question?
    \item \textbf{Conciseness:} Is the answer concise and clear, avoiding redundancy and repetition? A brief answer is important and will score higher.
    \item \textbf{Structure:} Is the answer well-organized and easy to understand quickly, with a clear focus on the main points?
    \item \textbf{Completeness:} Does the answer cover most of the key points related to the question, without omitting important information?
\end{enumerate}

We will provide the user's question, a high-quality reference answer, and the AI assistant's answer for evaluation. If the reference answer states that relevant content is not mentioned but the assistant's answer includes relevant information, the assistant’s answer should be rated higher, possibly achieving a perfect score of 10.

When you begin your evaluation, follow this process:
\begin{enumerate}
    \item Compare the AI assistant's answer with the reference answer, highlighting any shortcomings in the assistant’s response, and explain further. Keep in mind that the reference answer may not always be better than the assistant’s answer. The reference answer's quality level ranges from 3 to 10, so judge accordingly.
    \item Evaluate the AI assistant’s answer in different dimensions, giving a score between 1 and 10 for each.
    \item Finally, provide a comprehensive score based on the evaluations in each dimension.
    \item Your scoring must be as strict as possible, following these rules:
    \begin{itemize}
        \item Significant factual errors or irrelevant/harmful content should result in a low score (1-2).
        \item Answers with no serious errors but of low quality should receive a low score (3-4).
        \item Answers that generally meet the requirements but are weaker in some dimensions can receive a middle score (5-7).
        \item Answers that are similar to or slightly worse than the reference answer can receive a score of 8-9.
        \item Answers that surpass the reference answer by being more concise and nearly perfect in all dimensions can receive a score of 10.
    \end{itemize}
    Example: A reference answer can score 10 points.
\end{enumerate}

At the end of the evaluation, provide your scoring results in a dictionary format with the following keys:
\begin{quote}
\{'Fact Accuracy': X, 'User Needs Fulfillment': X, 'Conciseness': X, 'Structure': X, 'Completeness': X, 'Overall Score': X\}
\end{quote}
Remember to evaluate and explain each dimension and ensure that each score is an integer between 1 and 10.

Following examples show the evaluation process of CompassJudger:

\textbf{User's Question:} How can innovation improve the competitiveness of the real estate and renovation industries?

\textbf{Reference Answer:} 
{\color{olive}
Innovation can enhance the competitiveness of the real estate and renovation industries by utilizing technology such as big data and AI to optimize market analysis and customer demand prediction, developing eco-friendly materials to reduce environmental impact, implementing smart home technologies to improve quality of life, and offering personalized design services to meet customer needs.}

\textbf{AI Assistant's Answer:} 
{\color{teal}
Hello, Jiaojiao.

The real estate and renovation industries can innovate in several ways to improve their competitiveness:

1. Use of digital technology: Utilize technologies such as the internet, IoT, and big data to improve business efficiency, provide better customer experience, and reduce operational costs.

2. Promote green and eco-friendly concepts: With growing consumer awareness of environmental protection, the real estate and renovation industries can promote eco-friendly concepts and provide products and services that align with consumer demands.

3. Develop smart homes: Smart homes are a future trend. The real estate and renovation industries can preemptively develop related products and services to capture market opportunities.

4. Improve design quality: By collaborating with designers, the industry can enhance product design to offer more personalized and differentiated products and services.

5. Build brand reputation: Through offering high-quality products and services, businesses can build a strong brand reputation and attract more consumers.

These are some potential innovative strategies, and their implementation should be based on the specific situation of each company.}

\textbf{Evaluation Results:}
\begin{quote}
{\color{violet}
\{'Fact Accuracy': 9, 'User Needs Fulfillment': 8, 'Conciseness': 8, 'Structure': 9, 'Completeness': 8, 'Overall Score': 8\}}
\end{quote}
\subsection{Query 2: How should real estate companies adjust their strategic layout and break the homogeneous competition to adapt to market demand and urbanization development trends, while predicting and responding to the risks of a real estate bubble?}

\textbf{Reference Answer:} 
{\color{olive}Real estate companies should optimize their commercial real estate model, innovate business strategies, and break free from homogeneous competition. Simultaneously, market research should be strengthened to precisely target consumers and offer diversified products. Regarding the risk of a real estate bubble, companies should adopt measures like reasonable pricing and accelerating capital turnover to avoid crises.}

\textbf{AI Assistant's Answer:} 
Real estate companies should readjust their strategic planning, optimizing existing commercial real estate models to align with market demand and urbanization trends. To counter homogeneous competition, companies need to innovate their business models. Meanwhile, attention should be paid to the risks of a real estate bubble by taking measures such as reasonable pricing and accelerating capital turnover to mitigate potential crises. In addition, companies should strengthen market research, precisely target their customer base, and offer diversified products to meet various consumer demands.

\textbf{Evaluation Results:}
\begin{quote}
{\color{violet}
\{'Fact Accuracy': 9, 'User Needs Fulfillment': 8, 'Conciseness': 8, 'Structure': 9, 'Completeness': 8, 'Overall Score': 8\}}
\end{quote}

The response provided by the assistant successfully addresses the question by offering practical steps and techniques for enhancing sales and predicting market trends.

\subsection{Prometheus Evaluation of AI Assistant's Responses}
\subsubsection{Query 1: How can innovation improve the competitiveness of the real estate and renovation industries?}

\noindent \textbf{Instruction:} How can innovation improve the competitiveness of the real estate and renovation industries?

\noindent \textbf{Reference Answer:} {\color{olive}Innovation can enhance the competitiveness of the real estate and renovation industries by utilizing technology such as big data and AI to optimize market analysis and customer demand prediction, developing eco-friendly materials to reduce environmental impact, implementing smart home technologies to improve quality of life, and offering personalized design services to meet customer needs.}

\noindent \textbf{AI Assistant's Answer:} 
{\color{teal}
Hello, Jiaojiao.

The real estate and renovation industries can innovate in several ways to improve their competitiveness:

1. Use of digital technology: Utilize technologies such as the internet, IoT, and big data to improve business efficiency, provide better customer experience, and reduce operational costs.

2. Promote green and eco-friendly concepts: With growing consumer awareness of environmental protection, the real estate and renovation industries can promote eco-friendly concepts and provide products and services that align with consumer demands.

3. Develop smart homes: Smart homes are a future trend. The real estate and renovation industries can preemptively develop related products and services to capture market opportunities.

4. Improve design quality: By collaborating with designers, the industry can enhance product design to offer more personalized and differentiated products and services.

5. Build brand reputation: Through offering high-quality products and services, businesses can build a strong brand reputation and attract more consumers.

These are some potential innovative strategies, and their implementation should be based on the specific situation of each company.}

\noindent \textbf{Feedback:} The response provides a comprehensive approach to enhancing the competitiveness of the real estate and interior design industry. It effectively addresses the user's question by proposing various strategies, such as digitalization, green initiatives, smart home technology, design improvement, and brand reputation enhancement. Each point made is relevant and aligns well with the user's inquiry about innovative methods to improve market position. Furthermore, the structure of the response is clear, making it easy to follow and comprehend. The response demonstrates a good grasp of the topic, with no significant omissions. It offers practical solutions without unnecessary elaboration, adhering closely to the user's needs. In conclusion, the response is informative, relevant, and coherent, making it a solid answer to the user's question.

{\color{violet}\noindent \textbf{Score:} 5}

\subsubsection{Query 2: How should real estate companies adjust their strategic layout and break the homogeneous competition to adapt to market demand and urbanization development trends, while predicting and responding to the risks of a real estate bubble?}

\noindent \textbf{Instruction:} Based on our previous discussions, how should real estate companies adjust their strategic layout to break homogeneous competition, adapt to market demand and urbanization trends, and anticipate and respond to the risks of a real estate bubble?

\noindent \textbf{Reference Answer:} {\color{olive}Real estate companies should optimize their commercial real estate model, innovate business strategies, and break free from homogeneous competition. Simultaneously, market research should be strengthened to precisely target consumers and offer diversified products. Regarding the risk of a real estate bubble, companies should adopt measures like reasonable pricing and accelerating capital turnover to avoid crises.}

\noindent \textbf{AI Assistant's Answer:} 
{\color{teal}
Real estate companies should readjust their strategic planning, optimizing existing commercial real estate models to align with market demand and urbanization trends. To counter homogeneous competition, companies need to innovate their business models. Meanwhile, attention should be paid to the risks of a real estate bubble by taking measures such as reasonable pricing and accelerating capital turnover to mitigate potential crises. In addition, companies should strengthen market research, precisely target their customer base, and offer diversified products to meet various consumer demands.}

\noindent \textbf{Feedback:} The response effectively addresses the core issues discussed by focusing on strategic adjustments, innovation, market research, product diversity, and risk management. It offers clear, actionable steps for businesses to adapt to changing market demands and urbanization trends, while also anticipating potential risks from housing bubbles. Furthermore, the response presents these points in a structured and coherent manner, with a logical progression that enhances its clarity and usability. The information is presented in a concise and straightforward manner without repetition, which makes it easy to understand and implement. Lastly, the response covers all key aspects mentioned in the reference answer and demonstrates a comprehensive understanding of the topic at hand. Hence, the response fulfills all criteria set forth in the score rubric, as it exhibits high accuracy, relevance, simplicity, structure, and completeness in addressing the given instruction. So the overall score is 5.

{\color{violet}\noindent \textbf{Score:} 5}


\subsection{4 types of query dataset examples}
\textbf{cognitive\_example}

\begin{graybox}
Example 1 - Cognitive Low Complexity:

Question: "What ingredients were mentioned in the meeting?"

This is a Cognitive (category: cognitive) Low Complexity (complexity: low) question because it only 
requires direct extraction of explicitly mentioned ingredients from the dialogue.

Example 2 - Cognitive Medium Complexity:

Question: "What different opinions did participants have about ingredient preparation?"

This is a Cognitive (category: cognitive) Medium Complexity (complexity: medium) question because it requires synthesizing and comparing viewpoints from multiple speakers.

Example 3 - Cognitive High Complexity:

Question: "Based on the discussion, why did some participants prefer certain cooking methods?"

This is a Cognitive (category: cognitive) High Complexity (complexity: high) question because it requires analyzing decision-making reasons and involves inference and evaluation.
\end{graybox}

\textbf{context\_example}

\begin{graybox}
Example 1 - Context No Dependency:

Question: "What kitchen tools are available?"

This is a Context-Dependent (category: context) No Context (complexity: no) question because the answer can be obtained from a single statement without requiring surrounding context.

Example 2 - Context Recent Dependency:

Question: "Why did they change their approach to ingredient preparation?"

This is a Context-Dependent (category: context) Recent Context (complexity: recent) question because it requires understanding the immediate preceding discussion turns.

Example 3 - Context Long-Range Dependency:

Question: "How has the menu plan changed since the start of the discussion?"

This is a Context-Dependent (category: context) Long-Range Context (complexity: long) question because it requires tracking information throughout the entire meeting.
\end{graybox}
\clearpage
\textbf{domain\_example}

\begin{graybox}

Example 1 - Domain General Knowledge:

Question: "How many servings of food did they prepare?"

This is a Domain Knowledge (category: domain) General Knowledge (complexity: general) question because it requires no specialized knowledge to understand and answer.

Example 2 - Domain Basic Knowledge:

Question: "Why did they choose specific cooking temperatures and times?"

This is a Domain Knowledge (category: domain) Basic Knowledge (complexity: basic) question because it requires fundamental cooking knowledge.

Example 3 - Domain Expert Knowledge:

Question: "From a food safety perspective, what improvements could be made to their ingredient handling?"

This is a Domain Knowledge (category: domain) Expert Knowledge (complexity: expert) question because it requires professional food safety knowledge.

\end{graybox}

\textbf{task\_example}

\begin{graybox}

Example 1 - Task Simple Recording:

Question: "List all ingredients that need to be purchased."

This is a Task Execution (category: task) Simple Recording (complexity: simple) question because it only requires listing explicitly mentioned information.

Example 2 - Task Moderate Processing:

Question: "Summarize the preparation steps and timing for each dish."

This is a Task Execution (category: task) Moderate Processing (complexity: moderate) question because it requires organizing and synthesizing information.

Example 3 - Task Complex Action:

Question: "Based on the discussion, propose improvements to the current cooking process."

This is a Task Execution (category: task) Complex Action (complexity: complex) question because it requires analysis and strategic planning.

\end{graybox}
\clearpage
\subsection{Prompts}\label{app:Prompts}
We provide the prompts that we used to accomplish problem type routing, query the MeetMaster , create dataset and ground truth. 
\subsubsection{Planner query classifying prompt}
\textbf{query classifying prompts}
\begin{graybox}

    As a professional meeting assistant, please analyze the following question's type and complexity level.

\#\#\#  PROBLEM  \#\#\#

Who was asked to take over potato duty?

\#\#\#

Please classify strictly according to these standards:

1. Question Category (category) must be one of:

- cognitive: Questions requiring understanding and processing information

- context: Questions requiring consideration of dialogue context

- domain: Questions requiring specific domain knowledge

- task: Questions requiring task execution

2. Complexity Level (complexity) must be one of the following based on category:

For Cognitive questions:

- low: Simple recall/fact retrieval

- medium: Information synthesis/comparison

- high: Analysis/evaluation/inference

For Context-Dependent questions:

- no: No context dependency

- recent: Recent context dependency

- long: Long-range context dependency

- cross: Cross-meeting context dependency

For Domain Knowledge questions:

- general: General knowledge

- basic: Basic domain knowledge

- expert: Expert domain knowledge

For Task Execution questions:

- simple: Simple recording/listing

- moderate: Information organization/synthesis

- complex: Analysis/planning/strategy

\{cognitive\_example\}

\{context\_example\}

\{domain\_example\}

\{task\_example\}

If type is \textbf{Cognitive Load Low} or \textbf{Context Dependency None} or Domain Knowledge General385,first output "T!".
Please provide your response EXACTLY in the following JSON format, ensuring category and complexity values match the above list exactly:
\{\{
    "category": "cognitive",
    
    "complexity": "low",
    
    "explanation": "This is a cognitive low complexity question because..."
\}\}
\end{graybox}

\textbf{MeetMaster example output}

\begin{graybox}
\{
  "category": "context",
  
  "complexity": "recent",
  
  "explanation": "This is a context-dependent question with recent context dependency because it requires knowledge of who was assigned the task of handling potatoes, which would have been mentioned in the immediate discussion or meeting."
\}

\end{graybox}

\clearpage
\subsubsection{Planner three‑hop reasoning loop prompt}

\textbf{PLANNING\_PROMPT}

\begin{graybox}

<PLANNING\_PROMPT>

You are a conference assistant expert. The problem you need to solve is as follows. You can use the following tools to better solve the problem. You need to select a tool and output the input of the tool. Output according to the template.

\#\#\#  PROBLEM  \#\#\#

Who was asked to take over potato duty?

\#\#\#

\#\#\#  System tools available  \#\#\# 

• RagRetrieve      – company KB / meeting minutes  

• WebSearch        – open-web search  

• MeetSum          – abstractive segment summary
\#\#\#

\#\#\#  INPUT:  \#\#\#

\{

  "keywords": ["{k1}", "{k2}", …],

\}

\#\#\#

For each keyword:

  – choose the *most appropriate* tool
  
  – fetch the snippet ($\leq$ 60 words)  
  
  – record its source (url | “Meeting-MM/DD” | “KB-DocID”)
  
\#\#\#  TEMPLATE \#\#\#

Return a JSON list ($\leq$ 6 items):  

[

  {"tool": "RagRetrieve", "keywords": [LLM, Agent, …]},

]
\#\#\#

</PLANNING\_PROMPT>

\end{graybox}

\textbf{MeetMaster example output}

\begin{graybox}

\{

"tool": "RagRetrieve",

"keywords": ["potato duty", "take over"]

\}

\end{graybox}

\clearpage
\textbf{COMPOSE\_AND\_ACT\_PROMP}
\begin{graybox}

<COMPOSE\_AND\_ACT\_PROMPT>

You are a conference assistant expert.You are MeetMaster-Planner composing the final answer.The problem you need to solve is as follows.Your output according to the template

\#\#\#  INPUT:  \#\#\#

\{"query":  "{USER\_QUERY}",
  "evidence": [{tool, source, snippet}, …]   \# from Planning stage\}

\#\#\# 

\#\#\#  TASK  \#\#\# 

1  Write a concise, well-structured answer *in the user’s language*.  

2  Inline-cite evidence snippets as [1], [2] … (max 3).  

3  If appropriate, select an Action from the catalogue  

    – ScheduleMeeting, GenerateSlide, SendEmail, None  
    
    and fill its minimal JSON input.

\#\#\# 

\#\#\#  OUTPUT  (strict JSON, no extra text) \#\#\# 

\{"answer":  "<final answer with [x] citations>",  \}

\#\#\# 

</COMPOSE\_AND\_ACT\_PROMPT>

\end{graybox}

\textbf{MeetMaster example output}

\begin{graybox}

\{"answer":  "According to the result of RAG, P51 was asked to take over potato duty."  \}

\end{graybox}

\clearpage
\subsubsection{Creating Queries Dataset Prompt}
\textbf{Cognitive Complexity}
\begin{graybox}
I'm creating a dataset for meeting assistant agents that can answer questions of varying cognitive complexity. Given the following meeting transcript, please generate three questions (and their answers) that represent different cognitive complexity levels:

    1. Low complexity: Simple recall/fact retrieval questions about explicitly stated information
    
    2. Medium complexity: Questions requiring information synthesis or comparison across different parts of the discussion
    
    3. High complexity: Questions involving analysis, evaluation, or inference beyond what is explicitly stated

    MEETING TRANSCRIPT:
    
    {context}

    IMPORTANT: Please provide your response EXACTLY in the following format, with no additional text:

    Low complexity example:
    Question: "What was the increase in social media engagement mentioned by P02?"
    
    Ground truth answer: "P02 mentioned a 12\% increase in social media engagement after implementing the new content calendar."
    
    The complexity level: "Low complexity"
    
    Explanation of why it belongs to that complexity level: "This question is classified as low complexity because it requires simple recall of an explicitly stated fact from the transcript. The answer can be directly retrieved from P02's statement without requiring any synthesis, comparison, analysis, or inference. It only asks for a specific numerical value (12\% increase) that was clearly mentioned in the meeting."

    Medium complexity example:
    
    Question: "Could you compare the conversion rates between testimonial posts and overall social media engagement?"
    
    Ground truth answer: "Testimonial posts had a conversion rate of 4.2\%, which is higher than the overall social media conversion rate that increased from 2.3\% to 3.1\%."
    
    The complexity level: "Medium complexity"
    
    Explanation of why it belongs to that complexity level: "This question is classified as medium complexity because it requires synthesizing and comparing information from different parts of the discussion. The answer cannot be found in a single statement but requires gathering data about testimonial post conversion rates and overall social media conversion rates from different parts of the transcript, then making a comparison between these values. This involves connecting related pieces of information that were mentioned separately."

    High complexity example:
    Question: "Based on the discussion, which marketing channel appears to be performing better in terms of customer conversion, and what factors might explain this difference?"
    
    Ground truth answer: "Social media appears to be performing better for conversions, with testimonial posts reaching 4.2\% conversion compared to email's 4.8\% click-through rate (which is an earlier stage in the conversion funnel). This difference might be explained by the authenticity of user testimonials creating higher trust, the visual nature of social media content, or potentially different audience demographics between channels."
    
    The complexity level: "High complexity"
    
    Explanation of why it belongs to that complexity level: "This question is classified as high complexity because it requires analysis, evaluation, and inference beyond what is explicitly stated in the transcript. Answering requires not only comparing conversion metrics across different marketing channels, but also evaluating which channel is truly performing better (considering that click-through rates and conversion rates represent different stages of the funnel). Furthermore, it asks for possible explanatory factors that might not be explicitly mentioned in the discussion, requiring domain knowledge and inference about marketing principles to propose plausible reasons for the performance differences."

    For each question, provide:
    
    - The question
    
    - The ground truth answer
    
    - The complexity level
    
    - Explanation of why it belongs to that complexity level
    
\end{graybox}
\clearpage
\textbf{Context Dependency}
\begin{graybox}
I'm creating a dataset for meeting assistant agents that can handle questions with varying degrees of context dependency. Given the following meeting transcript, please generate four questions (and their answers) that represent different context dependency levels:

    1. No context: Questions answerable without relying on surrounding context
    
    2. Recent context: Questions requiring information from the immediately preceding discussion
    
    3. Long-range context: Questions requiring information from earlier parts of the meeting
    
    4. Cross-meeting context: Questions requiring information that connects to previously mentioned projects or future work

    MEETING TRANSCRIPT:
    
    \{context\}

    IMPORTANT: Please provide your response EXACTLY in the following format, with no additional text:

    No context example:
    
    Question: "What are the three database options being discussed?"
    
    Ground truth answer: "The three database options being discussed are an optimized MySQL configuration, migrating to PostgreSQL, or adopting MongoDB as a NoSQL solution."
    
    The context dependency level: "No context"
    
    Explanation of why it belongs to that context dependency level: "This question can be answered directly from the explicit information provided in the transcript without requiring knowledge of the discussion flow, timing, or relationships between statements. The three database options are clearly listed as standalone facts that do not depend on surrounding context to understand."

    Recent context example:
    
    Question: "What was the concern P07 raised about the database options?"
    
    Ground truth answer: "P07 raised concerns about the long-term maintenance costs of the different database options."
    
    The context dependency level: "Recent context"
    
    Explanation of why it belongs to that context dependency level: "This question requires understanding the immediately preceding statements in the discussion. To identify P07's specific concern, one needs to follow the recent flow of conversation and identify what P07 just mentioned. The concern is temporally localized to the most recent part of the discussion rather than requiring information from earlier parts or across different discussions."

    Long-range context example:
    
    Question: "How does the MongoDB option compare to the budget mentioned earlier in the meeting?"
    Ground truth answer: "The MongoDB option would cost around \$60,000, which exceeds the \$50,000 budget allocated for database improvements that was mentioned earlier in the meeting. P01 noted they might need to phase the implementation if they choose this option."
    The context dependency level: "Long-range context" 
    
    Explanation of why it belongs to that context dependency level: "This question requires connecting information from different, non-adjacent parts of the transcript. To answer it correctly, one must recall the budget amount mentioned much earlier in the meeting (\$50,000) and compare it with the cost of the MongoDB option (\$60,000) discussed later. This requires keeping track of information across the entire timeline of the meeting, not just recent exchanges."

    Cross-meeting context example:
    
    Question: "How does the current database discussion relate to the analytics dashboard project mentioned at the end?"
    
    Ground truth answer: "Whatever database solution they choose needs to integrate with the analytics dashboard project that P08's team is developing, as mentioned by P03 at the end of the discussion."
    The context dependency level: "Cross-meeting context"
    
    Explanation of why it belongs to that context dependency level: "This question requires connecting the current topic (database selection) with a separate project (analytics dashboard) that bridges to future work. It involves understanding how different projects or initiatives relate to each other across time boundaries, not just within the current meeting. The answer involves recognizing implications for future integration work and dependencies between projects that span beyond the immediate discussion."

    For each question, provide:
    
    - The question
    
    - The context dependency level
    
    - The ground truth answer
    
    - Explanation of why it belongs to that context dependency level
\end{graybox}

\textbf{Domain Knowledge}
\begin{graybox}
I'm creating a dataset for meeting assistant agents that can handle questions requiring different levels of domain expertise. Given the following specialized meeting transcript, please generate three questions (and their answers) that represent different knowledge complexity levels:

    1. General knowledge: Questions answerable with basic understanding, requiring no specialized knowledge
    
    2. Domain basics: Questions requiring fundamental understanding of the field's concepts and terminology
    
    3. Domain expertise: Questions requiring deep specialized knowledge, ability to interpret technical data, or make field-specific inferences

    MEETING TRANSCRIPT:
    
    \{context\}

    IMPORTANT: Please provide your response EXACTLY in the following format, with no additional text:

    General knowledge example:
    
    Question: "How many patients were involved in the clinical trial described in the meeting?"
    
    Ground truth answer: "347 patients with treatment-resistant depression were enrolled in the Phase 2b study of compound XR-27."
    
    The knowledge complexity level: "General knowledge"
    
    Explanation of why it belongs to that knowledge complexity level: "This question requires no specialized knowledge to understand or answer. Anyone who can comprehend basic English and numbers can identify the patient count mentioned in the transcript. No medical or clinical trial expertise is needed to recognize or interpret this straightforward factual information. The question simply asks for a numerical value that is explicitly stated."

    Domain basics example:
    
    Question: "What primary measure was used to assess efficacy in this clinical trial, and what was the result for the high-dose group?"
    
    Ground truth answer: "The MADRS (Montgomery-Åsberg Depression Rating Scale) was used to assess efficacy. The high-dose XR-27 group showed a 4.3 point greater reduction in MADRS scores compared to placebo, which was statistically significant with a p-value of 0.008."
    
    The knowledge complexity level: "Domain basics"
    
    Explanation of why it belongs to that knowledge complexity level: "This question requires fundamental understanding of clinical research concepts and terminology. To answer correctly, one needs to know what 'efficacy' means in a clinical context, understand what a rating scale is and why it's used to measure outcomes, and grasp basic statistical concepts like 'reduction in scores,' 'compared to placebo,' and 'p-value' indicating statistical significance. This represents entry-level domain knowledge that someone with basic training in clinical research would possess."
    Which domain knowledge it belongs to: "Clinical research and psychiatry"

    Domain expertise example:
    
    Question: "Based on the QTc prolongation data mentioned and the pharmacokinetic profile, what considerations should be made regarding cardiac monitoring for patients if XR-27 advances to Phase 3 trials?"
    
    Ground truth answer: "Given the 2\% incidence of QTc prolongation in high-dose patients (though none exceeded the 500ms threshold), combined with the 18-hour half-life supporting once-daily dosing, Phase 3 trials should include regular ECG monitoring, particularly during dose escalation. Patients with pre-existing cardiac conditions or who are taking other medications known to prolong QTc should be either excluded or monitored more frequently. The lack of significant CYP enzyme interactions suggests lower risk of unexpected QTc effects due to drug-drug interactions, but this should still be verified with dedicated cardiac safety studies."
    
    The knowledge complexity level: "Domain expertise"
    
    Explanation of why it belongs to that knowledge complexity level: "This question requires sophisticated specialized knowledge spanning multiple areas of pharmaceutical research and clinical cardiology. To answer correctly, one must understand cardiac electrophysiology (QTc intervals and their clinical significance), pharmacokinetics (drug half-life and its implications for dosing), drug-drug interactions (particularly CYP enzyme pathways), clinical trial design for cardiac safety monitoring, and regulatory considerations for cardiac risk assessment. The answer requires integrating complex technical information to make expert-level recommendations that would typically come from experienced clinical pharmacologists, cardiologists, or senior clinical research scientists."
    Which domain knowledge it belongs to: "Clinical pharmacology, cardiology, and pharmaceutical development"

    For each question, provide:
    
    - The question
    
    - The knowledge complexity level
    
    - The ground truth answer
    
    - which domain knowledge it belongs to
    
    - Explanation of why it belongs to that knowledge complexity level
    
\end{graybox}
\clearpage
\textbf{Task Execution}

\begin{graybox}
I'm creating a dataset for meeting assistant agents that can handle tasks of varying execution complexity. 
Given the following meeting transcript, please generate three questions/requests (and their responses) that represent different task execution complexity levels:

    1. Simple recording: Tasks requiring simple information retrieval or documentation
    
    2. Moderate processing: Tasks requiring synthesis, organization, or summarization of multiple information points
    
    3. Complex action: Tasks requiring analysis, recommendation, strategic planning, or decision support based on the meeting content

    MEETING TRANSCRIPT:
    
    \{context\}

    IMPORTANT: Please provide your response EXACTLY in the following format, with no additional text:

    Simple recording example:
    
    Question: "Can you list all the team members present in this meeting and their roles?"
    
    Ground truth answer: "The team members present are: P01 (Project Manager), P02 (IT Implementation Lead), P03 (HR Director), and P04 (Finance Manager)."
    
    The task execution complexity level: "Simple recording"
    
    Explanation of why it belongs to that task complexity level: "This task involves straightforward information retrieval without requiring any processing or transformation of the information. The assistant simply needs to identify and list explicitly stated facts from the transcript - the team members' names and their roles. No synthesis, analysis, or inference is needed; the information can be directly extracted and presented in the same form it appears in the meeting."

    Moderate processing example:
    
    Question: "Could you create a summary of the project timeline with all the adjusted dates mentioned in the discussion?"
    
    Ground truth answer: "Project Timeline:

    Requirements gathering: January (sign-off by January 31st)
    
    System configuration: February to March (completion by March 23rd, reduced from 10 to 8 weeks)
    
    Testing: April to mid-May (extended from 4 to 6 weeks, completion by May 15th)
    
    Training: Rest of May (completion by May 31st, with potential challenges due to P03's annual performance review cycle)
    
    Go-live: June 15th
    
    The entire implementation period spans 6 months."
    
    The task execution complexity level: "Moderate processing"
    
    Explanation of why it belongs to that task complexity level: "This task requires synthesizing and organizing information from multiple parts of the discussion into a coherent structure. The assistant needs to identify all timeline-related information scattered throughout the transcript, understand how dates were adjusted from original plans, organize this information chronologically, and present it in a structured format. This involves more than simple retrieval - it requires comprehension of how different timeline elements relate to each other and the creation of a new information structure (the timeline summary) that wasn't explicitly provided in the original discussion."

    Complex action example:
    
    Question: "Based on the concerns raised in the meeting, what are the top three risk factors for this project, and what mitigation strategies would you recommend?"
    
    Ground truth answer: "Top three risk factors and mitigation strategies:

    Timeline pressure: The testing phase was initially underestimated, and HR has conflicts during the training period. Mitigation: The team has already extended testing from 4 to 6 weeks, but should also consider creating a detailed testing plan early and potentially offering flexible training schedules to accommodate HR's performance review cycle.
    
    Budget management: The Finance Manager highlighted previous budget overruns (20\% on the ERP module). Mitigation: While there is currently a buffer (\$420,000 budget vs. \$395,000 proposal plus 15\% contingency), implementing strong change management procedures and regular budget reviews would help prevent scope creep.
    
    Integration challenges: The Finance Manager mentioned the need for proper integration with accounting systems. Mitigation: Schedule dedicated integration testing sessions with the Finance team, document all integration requirements thoroughly during the January requirements phase, and possibly involve integration specialists from the vendor."
    
    The task execution complexity level: "Complex action"
    
    Explanation of why it belongs to that task complexity level: "This task demands high-level analytical thinking and strategic recommendation development based on the meeting content. The assistant must: 1) analyze the entire discussion to identify implicit and explicit risk factors, 2) evaluate and prioritize these risks to determine the top three, 3) formulate specific mitigation strategies that aren't directly mentioned in the transcript, and 4) provide actionable recommendations that integrate project management best practices with the specific context of this project. This requires going beyond the literal content of the meeting to provide decision support that combines domain expertise with critical analysis of the situation."

    For each question/request, provide:
    
    - The question/request
    
    - The task execution complexity level
    
    - The ground truth response
    
    - Explanation of why it belongs to that task complexity level
\end{graybox}
